\documentclass[letterpaper, 10 pt, conference]{ieeeconf}
\usepackage{cite}
\usepackage{amsmath,amssymb,amsfonts,pifont}
\usepackage[ruled,linesnumbered]{algorithm2e}
\usepackage{algorithmic}
\usepackage{graphicx}
\usepackage{subfigure}
\usepackage{graphics}
\usepackage{textcomp}
\usepackage{xcolor}
\usepackage{mathptmx} % assumes new font selection scheme installed
\DeclareMathAlphabet{\mathcal}{OMS}{cmsy}{m}{n}
\DeclareSymbolFont{largesymbols}{OMX}{cmex}{m}{n}
\usepackage{times} % assumes new font selection scheme installed
\usepackage{bm}
\usepackage{multirow}
\usepackage{booktabs}
\usepackage{array}
\usepackage{tabularx,booktabs}
\usepackage{overpic}
\usepackage{url}
\usepackage[bookmarks=false]{hyperref}
\usepackage{rotating}
\usepackage{mathtools}
\usepackage{cite}
\usepackage{multirow}
\usepackage{balance}
\IEEEoverridecommandlockouts                              % This command is only needed if 
\usepackage{pifont}

\usepackage{verbatim}
\usepackage{CJKutf8}

\title{\LARGE \bf 
        Rendering-Enhanced Automatic Image-to-Point Cloud Registration for Roadside Scenes}
        
\author{Yu Sheng$^{1}$, Lu Zhang$^2$, Xingchen Li$^{1}$, Yifan Duan$^1$, Yanyong Zhang$^{1}$, Yu Zhang$^{1}$, and Jianmin Ji$^{1,2, \dag}$ 
\thanks{$^1$ School of Computer Science and Technology, University of Science and Technology of China (USTC), Hefei 230026, China}
\thanks{$^2$ Institute of Artificial Intelligence, Hefei Comprehensive National Science Center, Hefei, Anhui, China}
\thanks{ $^\dag$ Corresponding author. {\tt\small jianmin@ustc.edu.cn}}}
\begin{document}
\maketitle
\thispagestyle{empty}
\pagestyle{empty}

\begin{abstract}
% Prior point clouds offer 3D information of the background environment, which can assist roadside monocular cameras to estimate the depth of the objects in the scene after completing the registration between corresponding images and point clouds, thereby facilitate multiple downstream tasks, like 3D object detection, in roadside applications.
%assisting roadside cameras aiding roadside cameras in measuring distances in the real world, which facilitates many downstream tasks in roadside scenes. 
Prior point cloud provides 3D environmental context, which enhances the capabilities of monocular camera in downstream vision tasks, such as 3D object detection, via data fusion. 
However, the absence of accurate and automated registration methods for estimating camera extrinsic parameters in roadside scene point clouds notably constrains the potential applications of roadside cameras.
%precise automated registration methods between prior point clouds and roadside cameras limits the application in this scenario.
This paper proposes a novel approach for the automatic registration  between prior point clouds and images from roadside scenes.
%the roadside camera and prior point clouds.
The main idea involves rendering photorealistic grayscale views taken at specific perspectives from the prior point cloud with the help of their features like RGB or intensity values.
These generated views can reduce the modality differences between images and prior point clouds, thereby improve the robustness and accuracy of the registration results.
%with the core idea of synthesizing realistic images from point clouds with the pseudo-camera perspective to reduce modality differences and improve registration accuracy.
Particularly, we specify an efficient algorithm, named \emph{neighbor rendering}, for the rendering process.
Then we introduce a method for automatically estimating the initial guess using only rough guesses of camera's position.
At last, we propose a procedure for iteratively refining the extrinsic parameters by minimizing the reprojection error for line features extracted from both generated and camera images using Segment Anything Model (SAM).
We assess our method using a self-collected dataset, comprising eight cameras strategically positioned throughout the university campus.
Experiments demonstrate our method's capability to automatically align prior point cloud with roadside camera image, achieving a rotation accuracy of 0.202$^{\circ}$ and a translation precision of 0.079m.
Furthermore, we validate our approach's effectiveness in visual applications by substantially improving monocular 3D object detection performance.
% As far as we known, the approach is the first one that can automatically achieve accurate registration without a good guess of initial parameters. 
% Details of the experiments would be available online. 
%Specifically, we introduce an efficient algorithm to synthesize high-quality images from point clouds, dubbed \emph{neighbor rendering}.
%We propose a pipeline that automatically estimates initial parameters using only coarse camera positions, which significantly reduces the dependency of the registration algorithm on accurate initial parameters.
%Assisted by large visual model SAM, we precisely extract and match line features synthesized and camera images. 
%The camera pose is iteratively refined by minimizing the reprojection error associated with these line features.
%Extensive experiments on self-collected data demonstrate the effectiveness of our approach.

\end{abstract}
\section{Introduction}
\label{sec:intro}

Image-to-point cloud registration seeks to identify a rigid transformation for aligning  a 3D point cloud with a 2D image. 
This process, a cornerstone of multi-modal fusion, has attracted considerable attention in domains like autonomous driving and computer vision. 
Integrating prior point clouds with images has been proven to enhance multiple downstream tasks in roadside environments, such as vehicle speed measurements~\cite{vuong2024toward} and 3D object detection~\cite{ravi2018object_detection,ye2022rope3d}.
% This improvement is largely due to the precise geometric details—such as ground depth—offered by the  pre-existing point clouds.
% Despite significant advancements in automatic registration algorithms for vehicle-side and indoor scenes, the registration between roadside camera images and point clouds continues to depend largely on manual intervention.
Unlike the application fields of LiDAR-camera calibration~\cite{continue_line,koide2023general} and visual localization~\cite{cattaneo2019cmrnet,zhou2024differentiable}, the registration of roadside camera images with prior point clouds aims at finding the fixed cameras' poses to these point clouds, a task that is currently heavily reliant on manual techniques for its accomplishment~\cite{ye2022rope3d}.
% LiDAR-camera extrinsic calibration and visual localization represent the two main applications of image and point cloud registration.
% The distinction lies in the fact that, in visual localization, the point cloud is assumed to be pre-collected with the objective of estimating the pose of a moving camera within the point cloud.
% In contrast, in LiDAR-camera calibration, images and point clouds are collected simultaneously, with the goal of estimating the coordinate transformation relationship between the two sensors.
% Registration between roadside cameras images and point clouds is situated between these two applications, with the goal of determining the pose relationship of a fixed camera within a prior point cloud.

\begin{figure}[t]
\vspace{0.3cm}
\centering
\includegraphics[width=0.48\textwidth]{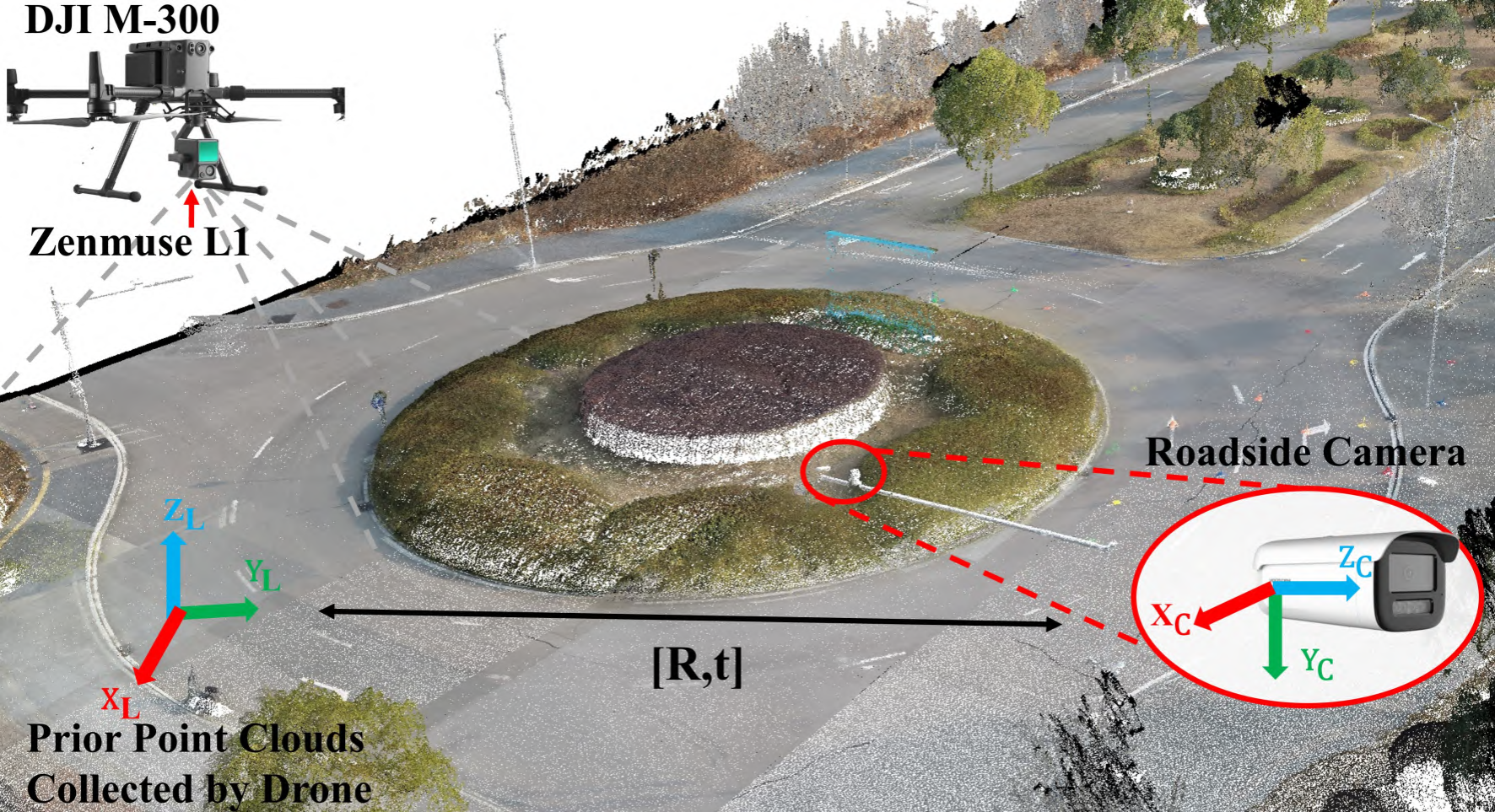}
\caption{Our method registers drone-collected point clouds with roadside camera images. This registration enhances monocular algorithm performance for downstream roadside tasks.}
% \begin{CJK*}{UTF8}{gbsn} 
% \lnote{这个roadside camera在图像中显示得实在是太不起眼了，很难发现，作为唯二的重要传感器，至少和无人机一样在图中能被一眼看到}
% \end{CJK*}
% (1) We check camera pose  based on its position and merge frames. (2) Initial guess are automatically detected through feature point matching. (3) An optimization process is conducted to refine the extrinsic parameters. The method requires  point cloud with RGB information, its corresponding images, and the estimated camera positions.}
\label{first}
\vspace{-1.0em}
\end{figure}

%Due to the characteristics of roadside scenes, applying
Given the distinct classifications of roadside environments, the direct usage of existing automatic registration methods
 may encounter a range of challenges. 
%Below, we point out a few of them.
Firstly, the prior point cloud and image are often captured at different time and within different coordinate systems. 
This results in a variation in the relative poses between different roadside cameras and point clouds, complicating the acquisition of precise initial guess needed by classical registration methods~\cite{continue_line,zhang2021line}.
While learning-based~\cite{LCCNet,cattaneo2019cmrnet} approaches have the potential to mitigate the dependency on initial guess through scene comprehension, acquiring a diverse and sufficient dataset for training is difficult in roadside settings. 
The recent work~\cite{koide2023general} manages to lessen the dependency of the registration algorithm on initial guess by transforming point clouds into images, thereby reducing the modal discrepancy between the two data.
However, \cite{koide2023general} struggles to address the unpredictable extrinsic parameters in roadside scenarios. 
Moreover, due to sparse and uneven distribution of roadside point clouds, directly projecting point clouds onto images as done in \cite{koide2023general,zhu2021camvox} can lead to issues such as the background being visible through the front surface (so-called bleeding problem), making it challenging to generate effective views for feature matching.
This constitutes the second challenge.
Although numerous methods, including NPBG~\cite{aliev2020neural} and 3DGS~\cite{kerbl20233dgs}, have been developed to generate new views from point clouds with remarkable results, directly applying these methods to roadside point clouds with a vast number of points would incur an unbearable computational cost.
Furthermore, these methods often overlook the 2D-3D correspondence during the rendering process, a key component necessary for accurate registration.
% \emph{3) Temporal asynchrony}.
% Point clouds and images are captured at different times, leading to changing local features within the scene, complicating the registration process.

To address the aforementioned challenges, we propose an automatic registration method for roadside camera images and prior point clouds.
We introduce \emph{neighbor rendering}, an efficient approach for generating realistic views from point cloud while preserved 2D-3D correspondence, ideal for registration.
% We propose an efficient rendering method suitable for registration, termed \emph{neighbor rendering}, to generate realistic views from point clouds while preserving 2D-3D correspondence.
Leveraging \emph{neighbor rendering}, we propose a initial guess estimation framework that involves pose sampling near rough camera location selected from point cloud and view generating, followed by utilizing SuperGlue~\cite{sarlin2020superglue} to establish 2D-3D.
% We sample key poses and synthesize views using \emph{neighbor rendering}. These synthesized views are then matched with camera images using SuperGlue~\cite{sarlin2020superglue} to refine the pose estimation range. Finally, we leverage the 2D-3D correspondence to extimate the initial guess.
Finally, we employ state-of-the-art (SOTA) segmentation  model SAM~\cite{kirillov2023segment} to extract line features from generated images and obtain corresponding  line features in the point cloud through the 2D-3D correspondence provided by \emph{neighbor rendering}.
The extrinsic parameters are optimized by minimizing the reprojection error.
We apply the registered image-point cloud pairs to roadside 3D object detection tasks, achieving remarkable performance and thus proving the effectiveness of our entire process.
It is noteworthy that, although our point clouds originate from DJI drones, our method seamlessly integrates with pre-existing maps produced by SLAM methods like R$^3$LIVE~\cite{R3LIVE}.
Our main contributions are as follows:
\begin{itemize}
    \item We propose a framework for automated image-to-point cloud registration in roadside scenes, utilizing view generated from point clouds to diminish data modality gaps, achieving SOTA performance in real-world roadside environments. 
    \item We introduce an efficient rendering method and utilize SAM to accurately extract edges from heterogeneous images, thereby enhancing registration accuracy.
    \item We apply our method to roadside 3D detection task to demonstrate the framework's practical effectiveness. 
\end{itemize}
\section{Related Work}
\label{sec:Related Work}
\subsection{Image-to-Point Cloud registration}

\subsubsection{Generic Image-to-Point Cloud registration}
% \begin{CJK*}{UTF8}{gbsn} 
% \lnote{建议引用使用超链接的格式，引用某个包就可以，可以方便查看}
% \end{CJK*}
Camera-LiDAR extrinsic calibration and visual localization are two key applications of image-to-point cloud registration. 
% The primary distinction lies in visual localization, where the point cloud is pre-collected, aiming to estimate the pose of a moving camera. In contrast, camera-LiDAR extrinsic calibration focuses on matching the image and point cloud by estimating the fixed relative pose between the two sensors.
% Our registration task aims to determine the fixed camera's pose within the prior point cloud, representing more of a fusion between the two.
% In the calibration tasks, cameras and LiDARs are co-located and collect data concurrently to establish the transformation relationship between the two sensors. 
Existing calibration methods between LiDAR and camera can be categorized into two types: target-based methods and target-less methods~\cite{li2022survey}.
Target-based methods typically use checkboards~\cite{zhang2004extrinsic} to establish the correspondence between LiDAR and images, while target-less methods~\cite{continue_line,zhang2021line} often relies on line or point features to achieve this objective.
\cite{koide2023general} optimize extrinsic parameters by minimizing the information discrepancy between LiDAR intensity images and grayscale images.
The same idea is applied in visual localization as well~\cite{pascoe2015direct}.
As neural network technology advances, \cite{feng20192d3d,ren2022corri2p,sun2022atop, LCCNet} leverages networks for identifying the correlations between 2D and 3D features, with subsequent refinement achieved through optimization with either the Perspective-n-Point (PnP) or Particle Swarm Optimization (PSO) algorithms. 
Another research trajectory~\cite{zhou2024differentiable} delves into the differentiability aspects of PnP, facilitating the end-to-end training of networks. 

\subsubsection{Roadside Image-to-Point Cloud registration}

In the camera-world calibration procedure for the Rope3D dataset~\cite{ye2022rope3d}, high-definition maps (HD Maps) containing lane and crosswalk endpoints are first projected onto 2D images to obtain the initial transformation, followed by bundle adjustment optimization.
\cite{jing2022intrinsic} utilizes manual matching of 2D-3D features followed by optimization using distance transform.
\cite{vuong2024toward} utilizes panoramic images for point cloud reconstruction and aligns roadside camera images with the reconstructed point cloud.
While these methods achieve excellent results, the need for manual intervention or obtaining panoramic images adds complexity to deployment.

\subsection{View Rendering}
\label{View Synthesi related}
Rendering views from point clouds is primarily achieved through two techniques: points-based rendering and meshing. 
Meshing facilitates the creation of high-quality images; however, it necessitates complex preprocessing step~\cite{kovavc2010point-based}.
The concept of utilizing points as rendering primitives was initially introduced by~\cite{levoy1985use,grossman1998point}, but it leads to bleeding problem.
Ray casting series~\cite{roth1982ray,hadwiger2005real} leverage surrounding points (volumes) to deduce the specific point on the ray, resulting in enhanced visual outcomes.
However, discerning spatial relationships in vast, unordered point clouds requires substantial computational cost.
\cite{pfister2000surfels} tackles these issues by substituting each point with oriented flat disks (a surfel), where the orientation and radius of these disks can be estimated from the point clouds.
This strategy underpins recent neural network-based rendering techniques such as  NPBG~\cite{aliev2020neural} and 3DGS~\cite{kerbl20233dgs}.
These methods focus more on generating high-quality views, thereby neglecting the 2D-3D correspondence needed by registration.
% Replacing points with surfel boosts rendering quality but increases computational demands. 
% Furthermore, the pixel-to-3D point correspondence, vital for registration, is lost.

\subsection{Roadside Camera Applications}
A thorough review of monocular roadside camera applications is presented in~\cite{Traffic_Review}.
Beyond tasks utilizing exclusively 2D data, like vehicle counting, applications requiring 3D information—for instance, vehicle speed measurement and vehicle distance estimation~\cite{Revaud_2021_ICCV,PlaneCalib,OptInOpt}—typically depend on the presumption of a flat ground or predefined size information. 
Such presumptions and prior knowledge considerably confine the algorithm's versatility across varied settings. 
Conversely, leveraging the 3D information from registered point cloud offers a robust solution to these limitations.
\section{Methodology}
\label{method}
\subsection{Overview}
\label{overview}

\begin{figure*}[t]
\vspace{0.3cm}
\centering
 \begin{minipage}{1\linewidth}
    \centering
        \includegraphics[width=1\linewidth]{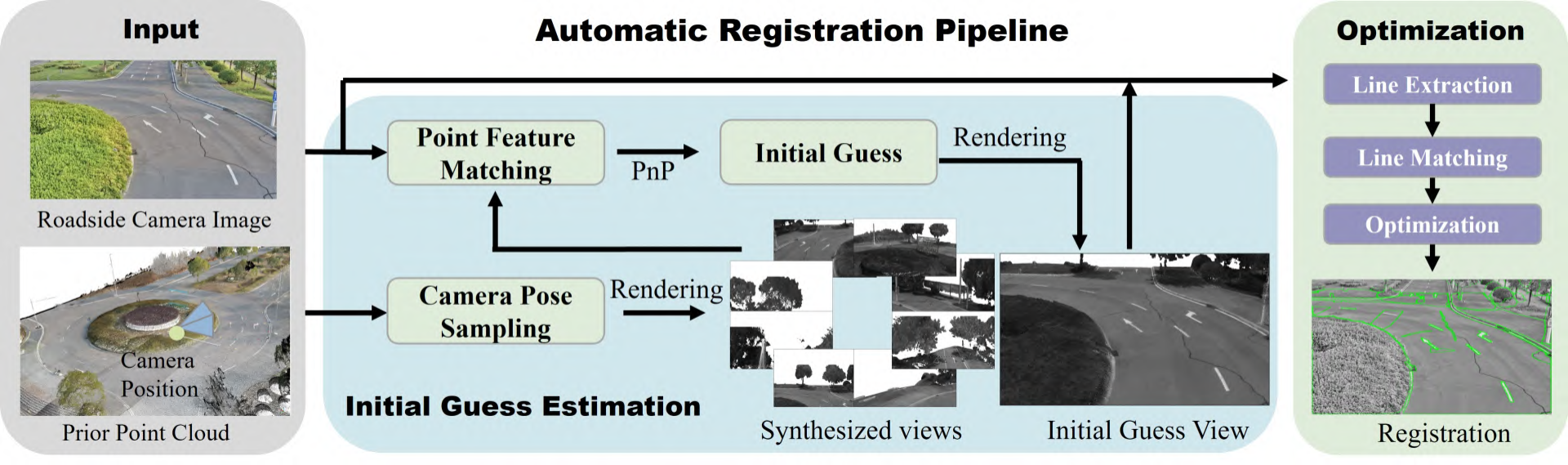}
    \end{minipage}
\caption{Our automatic registration method follows a two-step framework. (1) In initial guess estimation pipeline, we first sample camera poses around the rough location and synthesize views, then match this views with camera image using SuperGlue and estimate the initial guess by 2D-3D correspondence. (2) In optimization process, we first extract lines in point cloud and image, then we match these lines and optimize the extrinsics using reprojection errors.}
\label{pipeline}
\vspace{-1.0em}
\end{figure*}

\begin{figure}[t]
\vspace{0em}
\centering
\includegraphics[width=0.49\textwidth]{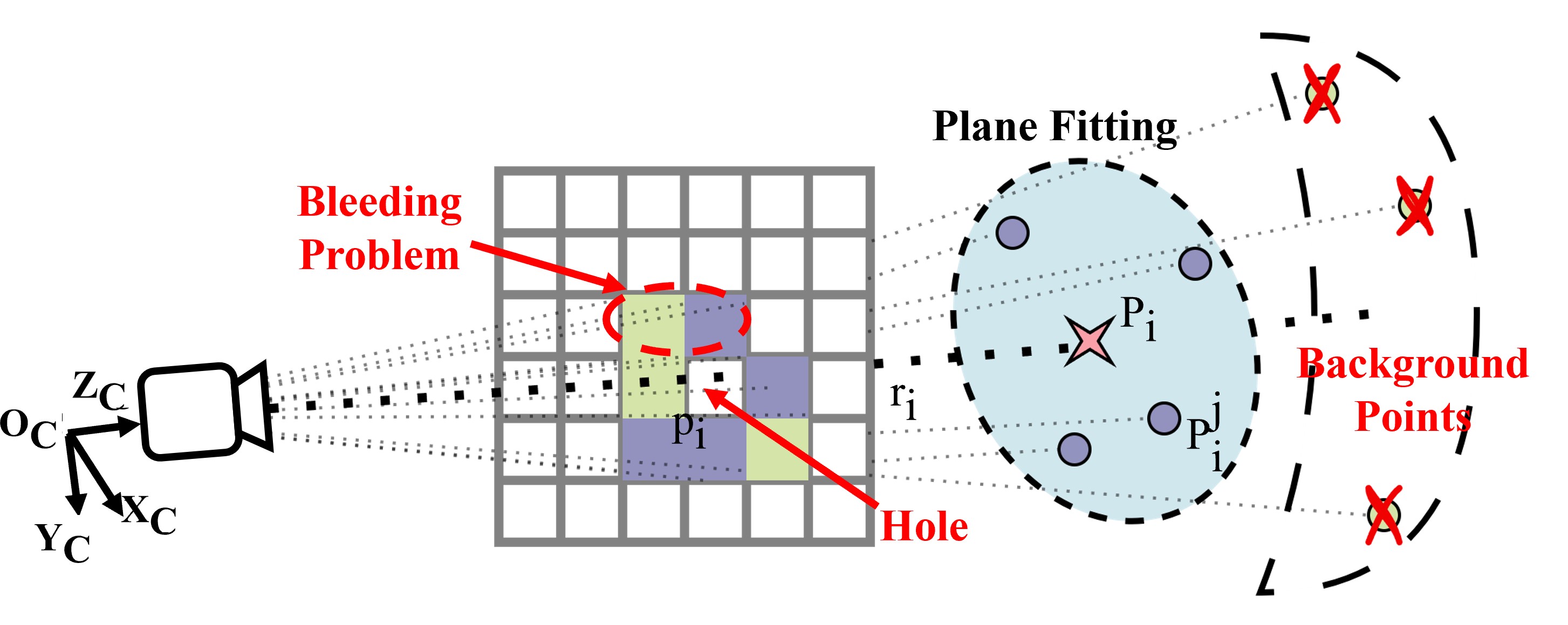}
\caption{
The purple points represent foreground points, while the green points represent background points. Neighbor rendering filters out the background points and estimate the 3D point $P_i$ corresponding to pixel $p_i$ based on the foreground points. }
% \begin{CJK*}{UTF8}{gbsn} 
% \lnote{这个caption太简略了，需要解释不同的颜色代表什么含义：哪个颜色的点代表背景点/需要过滤的，哪个需要保留？ 邻居点如何产生作用的？插图更容易传达方法的关键信息，每个图的caption都需要解释清楚}
% \end{CJK*}
\label{rendering}
\vspace{-1.0em}
\end{figure}

Given a pair of roadside camera image $\boldsymbol{\mathcal{I}} \in \mathbb{R}^{H*W*3}$ and a prior point cloud $\mathcal{P} \in \mathbb{R}^{N \times 3}$ with RGB values $\mathcal{C} \in \mathbb{R}^{N \times 3}$ or intensity $\mathcal{L} \in \mathbb{R}^{N \times 1}$ and rough camera position selected from point cloud, our method aims to estimate the grid transformation $T = [R,t] \in SE(3)$ between point cloud coordinates and camera coordinates, where $R \in SO(3)$ is the rotation matrix and $t \in \mathbb{R}^3$ is the translation vector.
% Here, we assume that the camera intrinsic parameters $K$ are known and the images have been undistorted.

In this section, we first detail the principles of \emph{neighbor rendering} as described in Section~\ref{Synthesis}. 
Following that, in Section~\ref{Initial Guess Estimating}, we propose a pipeline designed to obtain precise initial guess solely based on the selected rough camera position.
Lastly, in Section~\ref{Optimization}, we discuss optimizing extrinsic parameters by minimizing reprojection errors and effectively extracting line features using SAM.
The overview of our framework is shown in Fig.~\ref{pipeline}.

% We first place pseudo-cameras at different viewpoints around the rough location and synthesize views $I' \in \mathbb{R}^3$ using \emph{neighbor rendering}. 
% We match $I$ and $I'$ using SuperGlue until the number of matched feature points exceeds a threshold. 
% Subsequently, we estimate the initial camera extrinsic parameters $\Bar{T} = [\hat{R},\Bar{T}]$ using pixel-to-3D point correspondence and updating the pseudo-camera viewpoints with initial parameters.
% Finally, SAM~\cite{kirillov2023segment} is utilized to extract semantic line features between synthesized views and camera images, the extrinsic parameters is optimized by minimizing 2D-3D reprojection error.

\subsection{View Rendering}
\label{Synthesis}

% \begin{figure}[t]
% \vspace{0.2cm}
% \begin{center}
% 	\begin{minipage}{1\linewidth}
%         {\includegraphics[width=0.49\textwidth]{fig/bleeding.jpg}}
%         {\includegraphics[width=0.49\textwidth]{fig/point_size.jpg}}
% 	\end{minipage}
% 	\caption{Direct project points to images cause bleeding and poles (left), replace points with surfels can address this problem (right). Colored with intensity }
% \begin{CJK*}{UTF8}{gbsn} 
% \lnote{这个图传达效果一般，比较粗糙，如果时间充分可以考虑画示意图；
% 可以参考：Fig.1 from 
% LOG-LIO: A LiDAR-Inertial Odometry with Efficient Local Geometric
% Information Estimation 
% 以及：Fig.3 from 
% Robust Calibration of Vehicle Solid-State
% Lidar-Camera Perception System Using
% Line-Weighted Correspondences
% in Natural Environments 
% }
% \end{CJK*}
% \label{bleeding}
% \end{center}
% \vspace{-1.0em}
% \end{figure}

% As described in Eq.~\eqref{transformation}, A pixel $p_i(u_i,v_i)$ corresponds to a ray  $r_i(u_i,v_i) = t*K^{-1}[u_i,v_i,1]^T$ in the 3D space. 
% \begin{equation}
% r_i(u_i,v_i) = t*K^{-1}[u_i,v_i,1]^T
% \label{transformation}
% \end{equation}
Each pixel $p_i(u_i,v_i)$ corresponds to a ray  $r_i(u_i,v_i) = t*K^{-1}[u_i,v_i,1]^T$ in the camera coordinates. 
Without accounting for opacity, only the 3D point that falls on the ray and is closest to the camera can be projected onto the image.
However, due to the sparsity of roadside point clouds, lots of rays may not intersect with any 3D points, resulting in holes and bleeding problems (shown in Fig.~\ref{rendering}).
While previous studies like NPBG~\cite{aliev2020neural} and 3DGS~\cite{kerbl20233dgs} have made significant strides in addressing these issues, 
directly applying these methods to our task primarily faces two challenges: computational cost and the loss of 2D-3D correspondence, as detailed in Section~\ref{View Synthesi related}.
% Drawing insights from prior research, we propose an effective and efficient rendering method, termed \emph{neighbor rendering}, to tackle these challenges.
% \begin{CJK*}{UTF8}{gbsn} 
% \lnote{把这句话放在下一段的段首，更容易get到这一小节在做的事情}
% \end{CJK*}
% Below, we elaborate on the details of our approach.

Drawing insights from prior research, we propose an efficient rendering method, termed \emph{neighbor rendering}, to tackle these challenges.
Our method is based on two reasonable hypotheses:
1) When deduce the specific point on  a certain ray like~\cite{roth1982ray,hadwiger2005real}, only the 3D points surrounding this ray will be taken into consideration. 
2) Local 3d points can be approximated as belonging to a plane, a hypothesis also utilized in~\cite{pfister2000surfels}.
Specifically, we first project the point cloud onto the image and use the Z-buffer~\cite{aliev2020neural} to filter out occluded points.
% For pixels where no points are projected, we assign them sufficiently distant points.
This process not only filters occluded and out-of-view points but also reorganizes the point cloud, which enables retrieving points in the point cloud based on pixel coordinates, thus facilitating parallel processing.
For a given pixel $p_i$, we need to infer the points lying on its corresponding ray $r_i$.
Since adjacent 3D points are necessarily adjacent in pixel space, according to hypothesis~1, we only need to consider the 3D points retrieved from the $N$ neighbor pixels of $p_i$, denote as $\mathcal{Q}_i = \{P_i^j \in \mathbb{R}^3\mid j\in \{0,\ldots, N-1\}\,\}$.
% \dnote{It's a bit odd to use the same font for point clouds and points; I usually denote point clouds as $\mathcal{Q}$.}
Considering that these 3D points may not be adjacent in spatial, as shown in Fig.~\ref{rendering}, filtering out background points is necessary. 
We adopt a simple distance filter for background point filtering, as described in Eq.~\eqref{threshold}.
% only points with a distance smaller than 
% $\xi$ from the nearest point in $Q_i$ are considered as foreground points. 
Here, $d(\cdot)$ represents the distance from a 3D point to the camera and $\xi$ is a manually set threshold, $f(\cdot) = 1 $ indicates that the point is considered as a foreground point.
\begin{equation}
f(P_i^j) = \left\{
\begin{aligned}
	&1 \quad if \ d(P_i^j) - \min\{d(P_i^m)\mid P_i^m \in \mathcal{Q}_i\} \leq \xi,\\
	&0 \quad others.\\
	\end{aligned}
	\right.
\label{threshold}
\end{equation}

According to hypothesis~2, we fit plane function $a_ix+b_iy+c_iz+d_i = 0$ with foreground points and regard the intersection point of $r_i$ with this plane (denote as $P_i$) as the corresponding 3D point for $p_i$.
The appearance value of $P_i$ (denote as $v(P_i)$), such as the intensity or RGB values, are obtained by weighting the values of foreground points.
Given that points closer to $P_i$ should bear more resemblance to its appearance, inspired by the Inverse Distance Weighting (IDW)~\cite{itensity-generate}, we propose the Eq.~\eqref{contribution} to compute the weights $W(\cdot)$.
\begin{equation}
\begin{aligned}
v(P_i) &= \frac {\sum\limits^{N-1}\limits_{j=0,\,f(P_i^j)=1}W(P_i^j)\,v(P_i^j)}{\sum\limits^{N-1}\limits_{j=0,\,f(P_i^j)=1}W(P_i^j)}, \\
W(P_i^j) &= \frac {\xi + min\{d(P_i^m)\mid P_i^m \in \mathcal{Q}_i\} - d(P_i^j)}{\exp(D(P_i^j))},
\end{aligned}
\label{contribution}
\end{equation}
where $D(P_i^j)$ represents the distance of $P_i^j$ to the $P_i$. 
This equation indicates that points closer to $P_i$ have larger weights, and these weights decay exponentially as the distance increases.

% In practice, we discovered the impact of approximating the intersection point with the point closest to the ray and camera is minimal, yet it significantly boosts parallel processing speed.
We deploy our rendering algorithm using CUDA, allowing us to generate a 4K image in 20ms on RTX-3050 GPU.
The visualization of generated images are available in the appendix.

\subsection{Initial Guess Estimation}
\label{Initial Guess Estimating}

\begin{figure}[t]
\vspace{0.2cm}
\begin{center}
	\begin{minipage}{1\linewidth}
        {\includegraphics[width=0.45\textwidth]{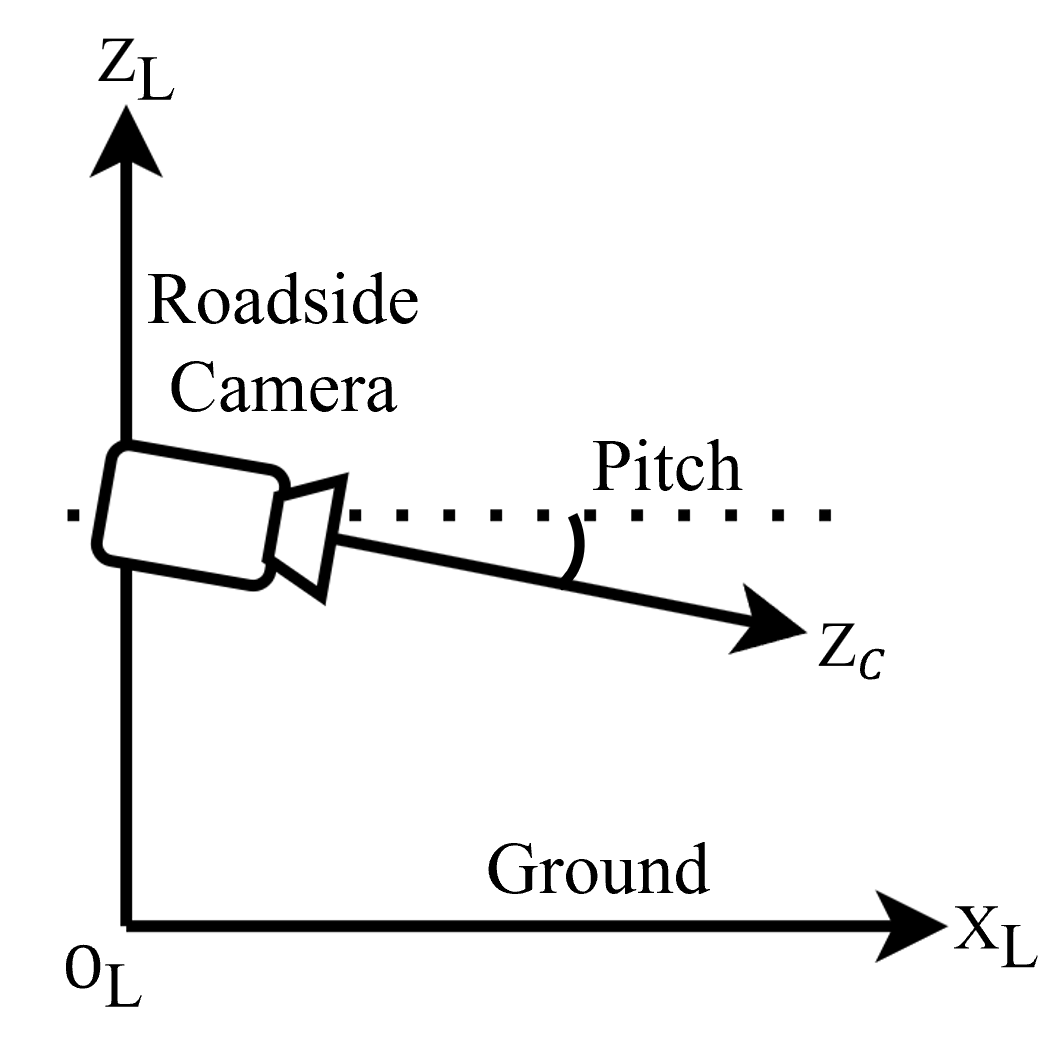}}
        {\includegraphics[width=0.45\textwidth]{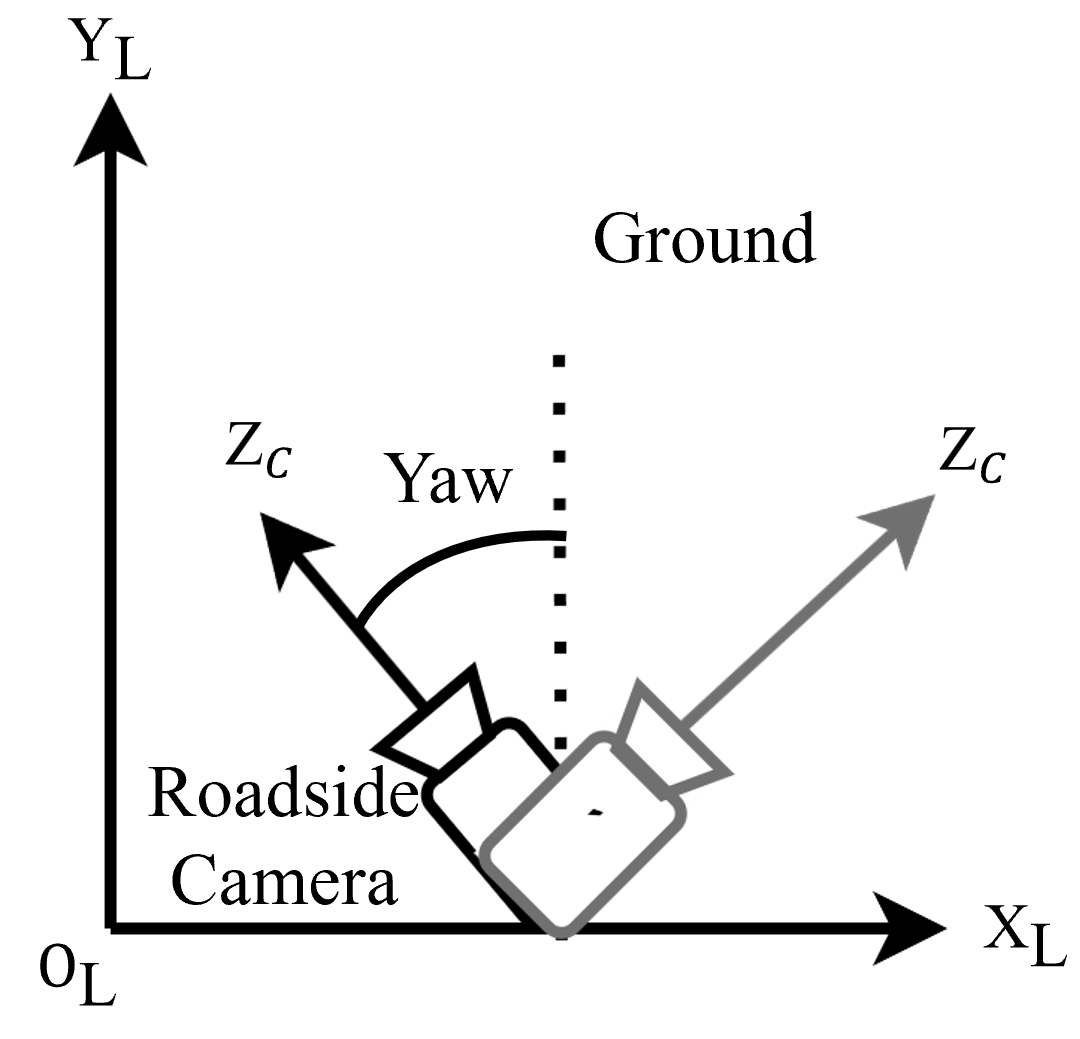}}
	\end{minipage}
	\caption{The typical installation for roadside cameras. We assume that the camera has a small pitch angle and place pseudo-cameras at different yaw angles to generate views.}
% \begin{CJK*}{UTF8}{gbsn} 
% \lnote{这个配图想要说明什么？
% 要不caption给的更详细，要不修改图直到与roadside相机的初始估计角度估计困难问题产生明显的联系}
% \end{CJK*}
\label{camera_fix}
\end{center}
\vspace{-2.0em}
\end{figure}

Through the analysis of typical installation configurations for roadside cameras, we observe that these cameras are characterized by an essentially zero roll angle and a slight downward pitch angle, which facilitates the observation of the ground area.
The yaw angle, conversely, is determined by the specific direction the camera is aimed at, as demonstrated in Fig.~\ref{camera_fix}. 
Building on this observation, we assign the camera a zero roll angle and a minor pitch angle. 
Within the expanse of the panoramic view, eight yaw angles are uniformly sampled and their respective views are generated through \emph{neighbor rendering}.
We then match these generated views with the camera image using SuperGlue~\cite{sarlin2020superglue} until a sufficient number of matched feature points obtained. 
Leveraging the 2D-3D correspondence preserved by \emph{neighbor rendering}, we utilize PnP~\cite{epnp} to estimate the rough pose of the camera.

Sampling merely eight yaw directions might result in partial matches between generated and camera images, leading to the clustering of matched feature points in particular regions. 
As indicated in~\cite{continue_line}, this uneven distribution of matched features heightens the uncertainty in extrinsic parameters estimation, making the estimation sensitive to environmental changes.
Thus, by regenerating view with the roughly estimated pose and matching again, we ensure that the initial guess are estimated under the condition of a uniform distribution of matched feature points.
% Therefore, we place a pseudo-camera at the estimated rough pose and synthesize views once more for matching. 
% This ensures an extensive spread of feature points across the image, facilitating a more accurate and stable initial guess.

\subsection{Extrinsic Parameters Optimization}
\label{Optimization}

\subsubsection{Line Extraction}
Edge extraction from images is well-established, yet deriving edges from point clouds poses greater difficulties. While depth discontinuity~\cite{zhang2021line} is a prevalent technique for edge detection in point cloud, it tends to yield jagged, imprecise edges. \cite{continue_line} pioneered the use of plane intersections for precise edge extraction from point clouds, showing notable efficacy. However, roadside scenes often lack sufficient planar features to meet the algorithm's requirements.
Furthermore, line features in roadside images primarily focus on surface textures like lane markings, which the aforementioned methods cannot directly extract.
Given our capability to generate realistic views and establish 2D-3D correspondences, it is possible to extract 2D line features from the generated images and then derive their corresponding 3D line features. 
However, traditional edge detection algorithm~\cite{canny1986computational} necessitate parameter tuning for optimal extraction, yet finding a unified setting for these parameters across heterogeneous image is difficult.
Therefore, we leverage the robust generalization capabilities of the powerful vision model SAM~\cite{kirillov2023segment} to facilitate precise extraction of line features in both generated and camera images.

\subsubsection{Optimization}
To align point cloud edges with those in images, we uniformly sample multiple points $P_L^i$ from each edge in the point cloud.  These points are then projected onto the image using current extrinsic estimate $\Bar{T}$ and camera intrinsic parameters $K$. 
% We assume images have been undistorted.
\begin{equation}
p_i = K\cdot\Bar{T}(P_L^i).
\label{projection}
\end{equation}
For each $p_i$, we search for its  $M$ nearest pixels $\mathcal{S}_i = \{ q_i^j \in  \mathbb{R}^2 \mid j\in \{0,\ldots,M-1\}\, \}$ in the $k$-D tree formed by image edge pixels.
The line formed by $\mathcal{S}_i$ represents the image edge corresponding to $p_i$.
To simplify computation, we parameterize the line as $q_i^0$ with the norm vector $n_i$, which is derived from the minimum eigenvector obtained through  through principal component analysis on $\mathcal{S}_i$ .

The extrinsic parameters are optimized by minimizing the projection error between $p_i$ and $q_i^0$ along $n_i$, which can be formulated as follows:
\begin{equation}
\begin{split}
    T^* &= \arg \underset{T} {\min} \sum\limits^{n}\limits_{i=1} \left(|| n_i^T(K\cdot T(P^L_i)- q_i^0)||^2\right) \\&= \arg \underset{R,t} {\min} \sum\limits^{n}\limits_{i=1} \left(|| n_i^T((K\cdot R(P^L_i)+t)- q_i^0)||^2\right).
\end{split}
\label{target_equ}
\end{equation}
To address the constraints introduced by the rotation matrix, we can utilize the transformation between Lie groups and Lie algebras:
\begin{equation}
\begin{split}
R &= E(\theta) = \exp(\theta^\wedge)\\ & = I + \sin(\theta)\theta^\wedge + (1-\cos(\theta))\theta^{\wedge 2}.
\end{split}
\label{lie algebras}
\end{equation}
where $\theta$ represents the Lie algebraic representation of the rotation matrix, and $\theta^\wedge$ denotes the corresponding skew-symmetric matrix. 
By transforming the optimization problem to the Lie algebra, the number of parameters to be optimized reduces to six, and the constraints are eliminated.
The optimization process can be solved recursively. 
Let $\Bar{T}$ be the current extrinsic estimate, using the operator $\boxplus $ encapsulated in the Special Euclidean group $SE(3)$~\cite{mainfold}, we can parameterize the target extrinsic $T$ as:
\begin{equation}
\begin{aligned}
T \; &\leftarrow \; \Bar{T} \boxplus_{SE(3)} \delta T \triangleq \exp(\delta T)\cdot \Bar{T}, \\ 
\delta T &= (\delta \theta, \delta t) \in \mathbb{R}^6.\\
\end{aligned}
\label{iteratively}
\end{equation}
Given the objective function defined in Eq.~\eqref{target_equ} and the operator $\boxplus $, we can employ ceres~\cite{Agarwal_Ceres_Solver_2022}  to efficiently solve the problem.
% Ceres~\cite{Agarwal_Ceres_Solver_2022}, an optimization library, can be employed to efficiently solve the problem in iterative way with the optimization objective function defined in Eq.~\eqref{target_equ} and the operator $\boxplus $ encapsulated in the Special Euclidean group $SE(3)$~\cite{mainfold}  as Eq.~\eqref{iteratively} illustrated. 

\section{Experiments}
% In this section, we evaluated our algorithm across real-world roadside scenarios dataset. 
\subsection{Dataset Preparation}
\begin{figure}[t]
\vspace{0.2cm}
\begin{center}
	\begin{minipage}{1\linewidth}
        {\includegraphics[width=0.49\textwidth]{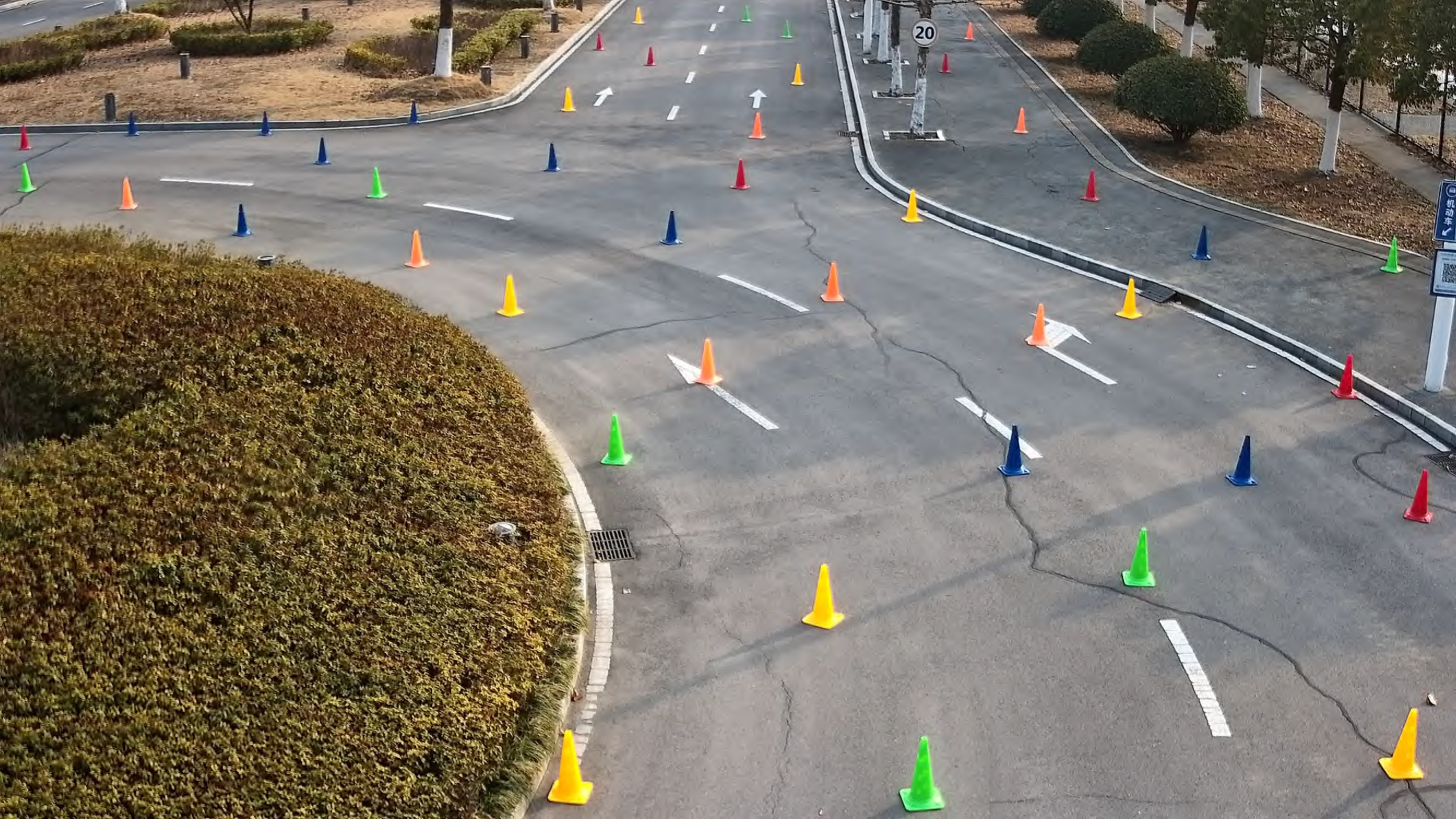}}
        {\includegraphics[width=0.49\textwidth]{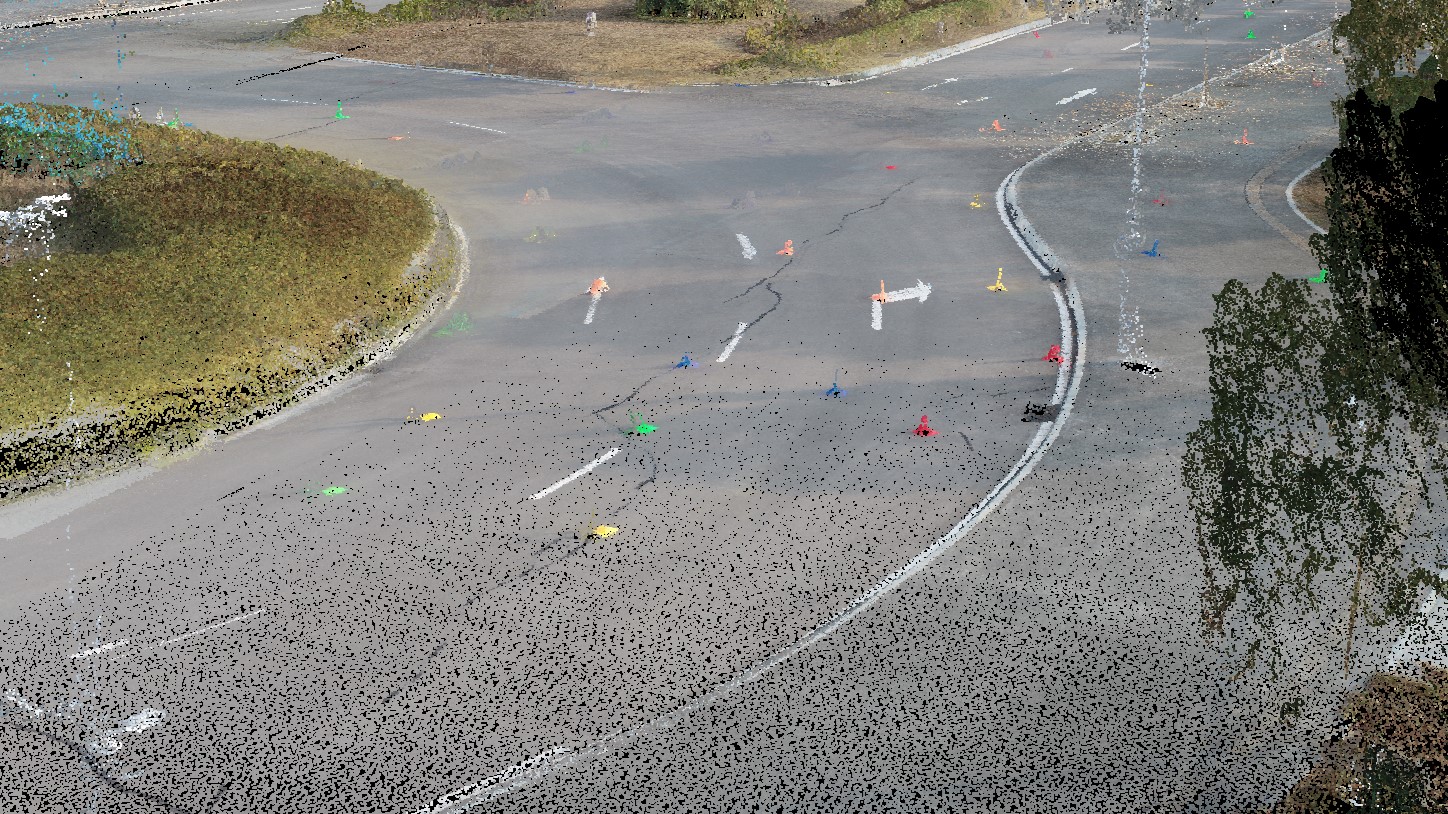}}
	\end{minipage}
	\caption{Left is cones in camera images, right is the corresponding point clouds, the points size has been enlarged for better visualization.}
\label{exp_set}
\end{center}
\vspace{-1.0em}
\end{figure}

We validate our method on a self-collected dataset. 
As shown in Fig.~\ref{first}, the prior point cloud is acquired using a DJI-M300 drone, equipped with the Zenmuse L1 module, alongside  Hikvision cameras mounted on roadside poles for image capture. 
The cameras' intrinsic parameters are calibrated using a checkerboard method~\cite{checkboard}. 
Data are gathered from eight unique areas, each providing two pairs of point cloud-image datasets, and all point clouds are aligned within a consistent global coordinate system. 
For each area,  one pair of point cloud-image is designated for evaluation, while the other include cones evenly distributed within the camera's field of view (FoV) to facilitate the acquisition of pseudo ground truth,  as shown in Fig.~\ref{exp_set}. 

To obtain the ground truth of the extrinsic parameters, We manually annotated 2D-3D corresponding points of clearly visible cones and minimized the reprojection error. 
The accuracy of this transformations are validated by projecting the point clouds onto the images, which are then used as pseudo ground truth.

% We uniformly placed cones within the camera's field of view (FoV) and captured the corresponding point clouds. Then we manually annotated 2D-3D corresponding points and minimized the reprojection error to obtain the transformation. 
% The accuracy of this transformation was validated by projecting the point clouds onto the images, and then adopted as pseudo ground truth.

% We deployed cameras on poles of eight distinct areas within campus and calibrated their intrinsic parameters using a checkerboard method~\cite{checkboard}.
% The acquisition of point clouds was conducted using the DJI MATRICE-300 drone, equipped with the Zenmuse L1 module, ensuring high-quality aerial point cloud data. 
% The point clouds collected from different areas were consistently aligned to the same UTM coordinate.
% The experimental configuration and details are shown in Fig.~\ref{first}. 
% For each scene, we collected two pairs of point clouds and image: one for validation analysis and the other containing cones in the scene, as shown in Fig.~\ref{exp_set}, used to derive pseudo ground truth of extrinsic parameters.

\begin{figure}[t]
\vspace{0.2cm}
\begin{center}
	\begin{minipage}{1\linewidth}
        {\includegraphics[width=0.32\textwidth]{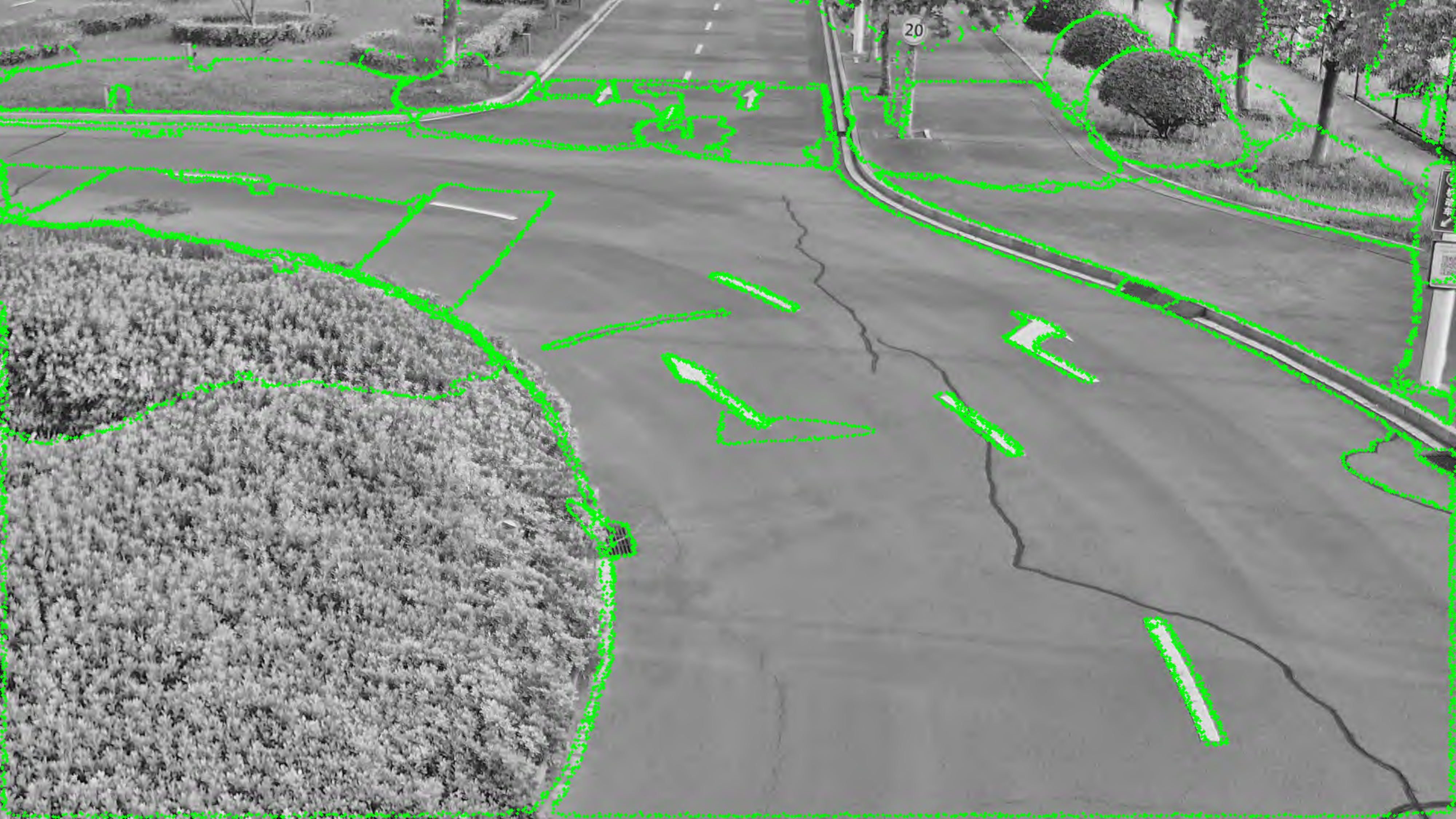}}
        {\includegraphics[width=0.32\textwidth]{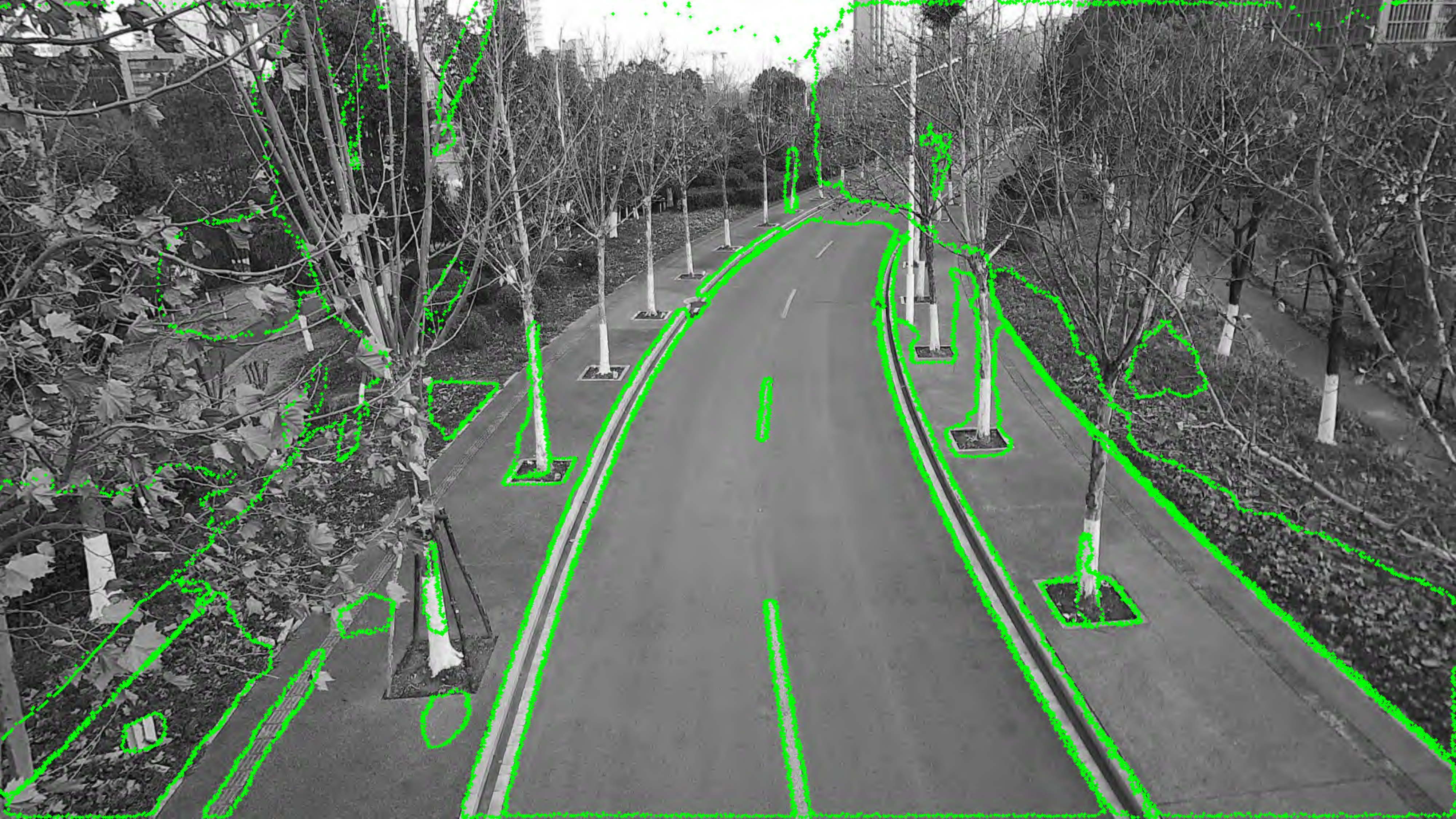}}
        {\includegraphics[width=0.32\textwidth]{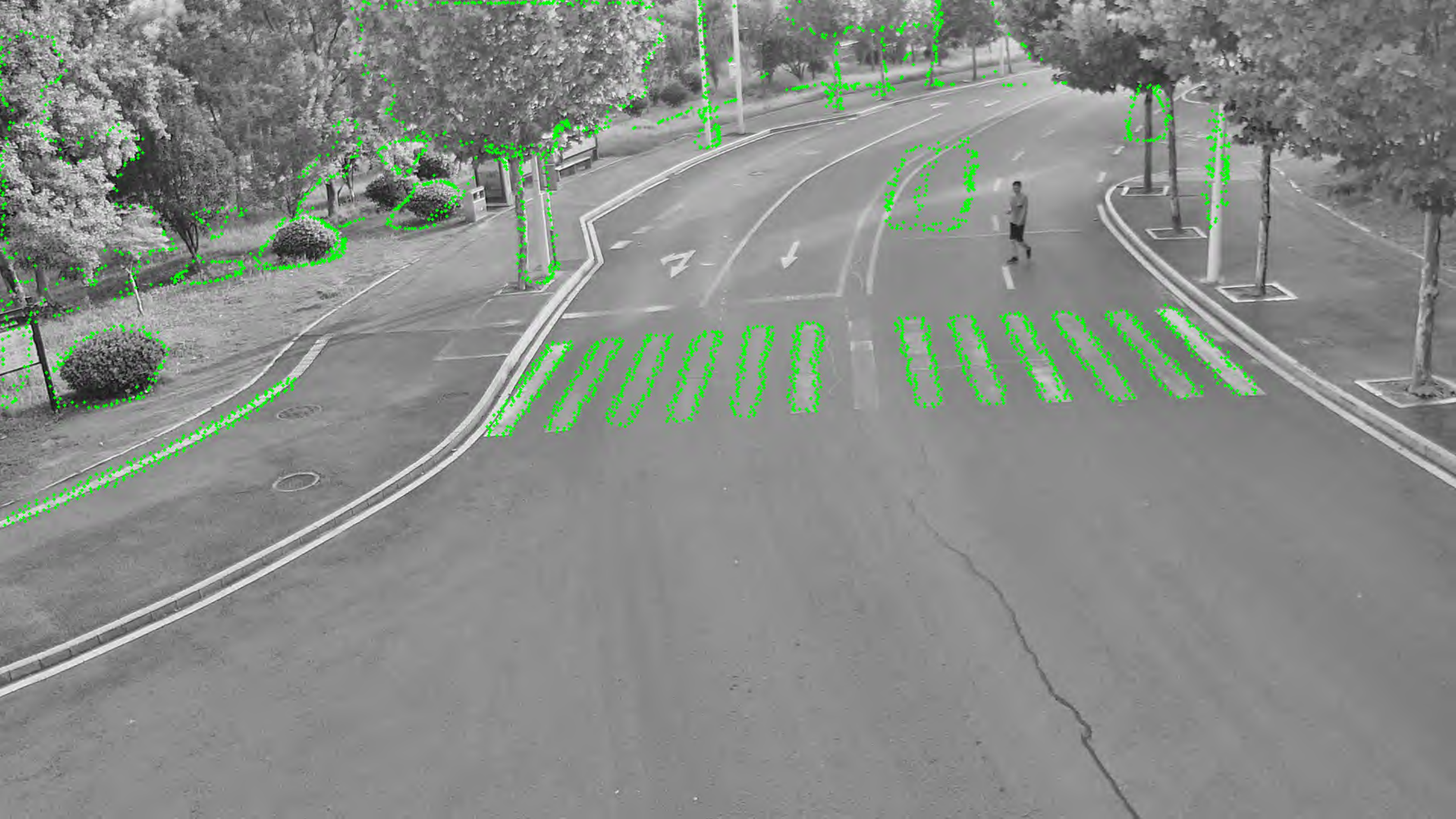}}
	\end{minipage}
	\caption{The qualitative registration results. The green lines represent the lines in the point cloud projected onto the camera plane using the estimated extrinsics. Visualization results for all scenarios can be found in the appendix.}
\label{qualitative}
\end{center}
\vspace{-2.0em}
\end{figure}

\begin{table}[t]
\vspace{1em}
    \setlength{\abovecaptionskip}{0em}
    \centering
    \caption{Quantitative Registration Results on Dataset}
    \label{registration_errors}
    \resizebox{\columnwidth}{!}{
    \begin{tabular}{c|cc|c|cc}
        \bottomrule[1.5pt]
        \  & \multicolumn{3}{c}{Ours} & \multicolumn{2}{c}{Continue Edge\cite{continue_line}} \\
        \  Scenes & Trans.(m)$\downarrow$ &  Rot.($^{\circ}$) $\downarrow$ & Init. Guess& Trans.(m)$\downarrow$ & Rot.($^{\circ}$)$\downarrow$\\
        \hline
        1 & 0.095 & 0.219 &\ding{51} & 0.157 & 0.375\\
        2 & 0.094 &  0.128 &\ding{51} & 0.684 & 0.487 \\
        3 & 0.045 &  0.102 &\ding{51} & 0.462 & 0.193 \\
        4 & 0.108 & 0.273 &\ding{51} & 0.709 & 1.237\\
        5 & 0.162 & 0.302 &\ding{51} & 1.098 & 1.455\\
        6 & 0.099 & 0.189 &\ding{51} & 0.416 & 0.601\\
        7 & 0.027 & 0.196 &\ding{51} & 0.088 & 0.270\\
        8 & 0.006 & 0.207 &\ding{51} & 0.096 & 0.390\\
        \toprule[1.5pt]
        % \pixel : 9.53 ,9.21,4.40, 10.79 , 7.37 ,11.15, 4.11, 6.73
        Avg. & \textbf{0.079} & \textbf{0.202} & 8/8 & 0.464 & 0.602
    % \multicolumn{4}{l}{$^{\mathrm{a}}$Sample of a Table footnote.}
    \end{tabular}}
\end{table}

\begin{table}[t]
\vspace{-1.0em}
    \setlength{\abovecaptionskip}{-0em}
    \centering
    \caption{Ground Distance Evaluation on Dataset}
    \label{Distance Evaluation}
    \resizebox{\columnwidth}{!}{
    \begin{tabular}{c|ccc}
        \bottomrule[1.5pt]
        \textbf{Methods} & Max Error($\%$)$\downarrow$ &  Median Error($\%$)$\downarrow$ &  RMSE($\%$)$\downarrow$ \\
        \hline
        OptInOpt~\cite{OptInOpt} &15.78& 10.80& 12.87\\
        PlaneCalib~\cite{PlaneCalib} &  14.32 & 9.23 & 11.69\\
        DeepVPCalib~\cite{DeepVP} &  12.17 & 8.11 & 10.62\\
    Revaud et al.~\cite{Revaud_2021_ICCV}& 14.87 & 10.91 &  12.54\\
        Vuong et al.~\cite{vuong2024toward} & 6.75 & 3.22 & 4.68 \\
        \toprule[1.5pt]
        \textbf{Ours} & \textbf{4.75} & \textbf{1.16} & \textbf{1.67} \\
    % \multicolumn{4}{l}{$^{\mathrm{*}}$Sample of a Table footnote.}
    \end{tabular}}
\vspace{-1.0em}
\end{table}

\begin{figure}[t]
\vspace{0.2cm}
\begin{center}
	\begin{minipage}{1\linewidth}
        {\includegraphics[width=0.49\textwidth]{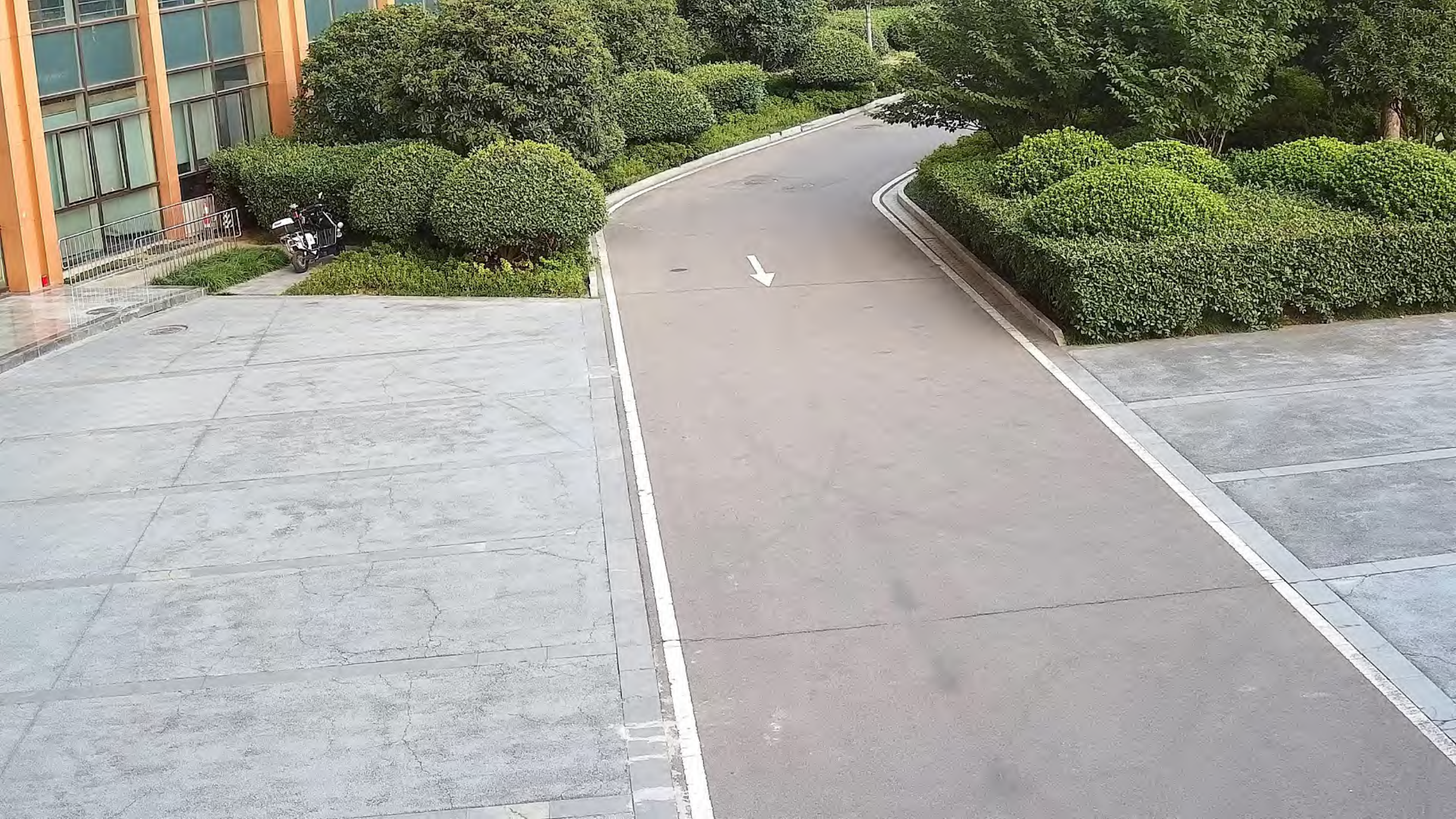}}
        {\includegraphics[width=0.49\textwidth]{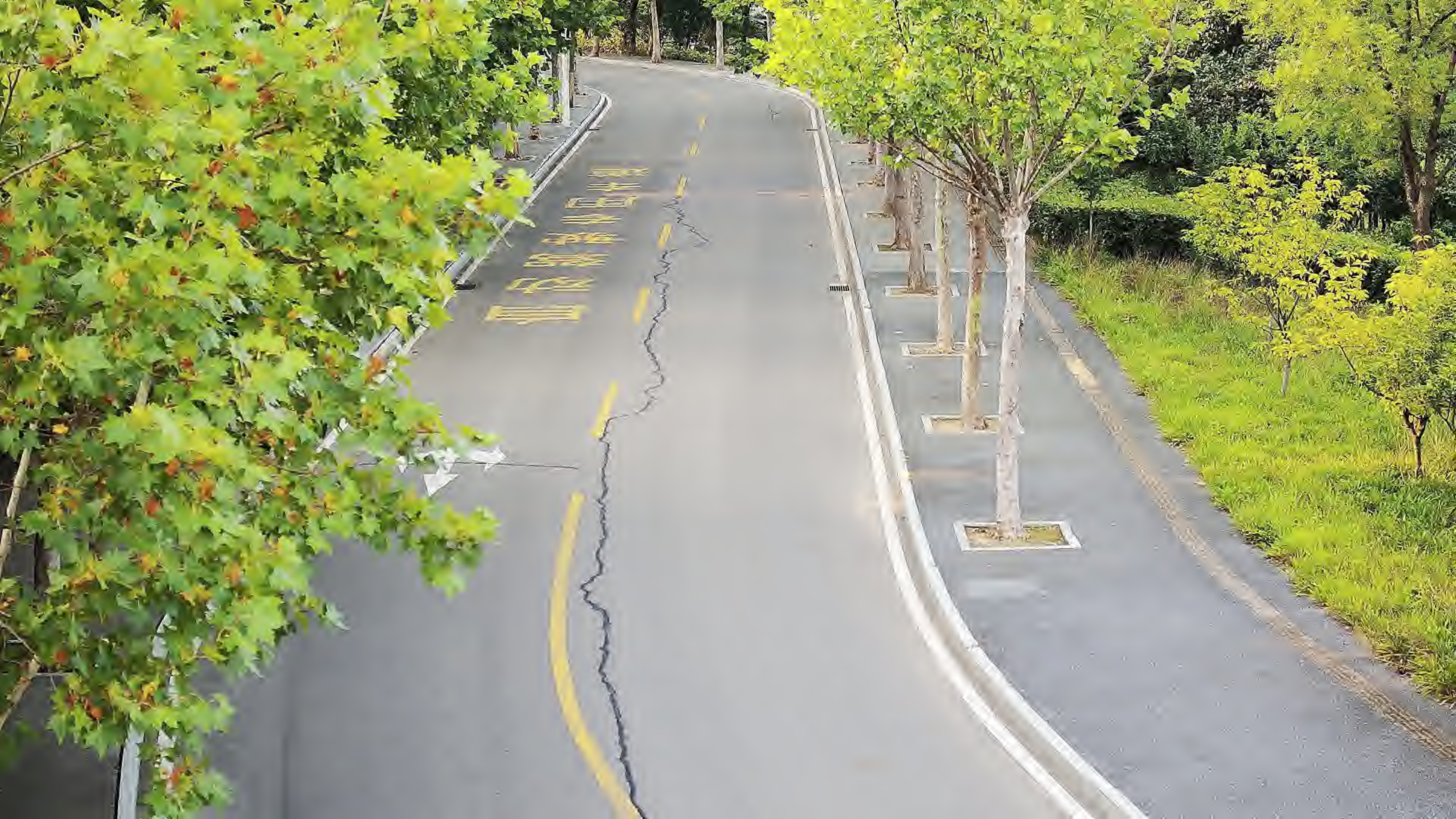}}
	\end{minipage}
	\caption{Examples of bad scenes. The scene on the left lacks sufficient features, while on the right, dense foliage obstructs the drone's ability to capture the point cloud.}
\label{bad scenes}
\end{center}
\vspace{-1.0em}
\end{figure}

% \begin{table}[t]
% \vspace{1.0em}
%     \setlength{\abovecaptionskip}{-0em}
%     \centering
%     \caption{Ablation Experiments Results}
%     \label{Ablation}
%     \resizebox{\columnwidth}{!}{
%     \begin{tabular}{ccc|cc}
%         \bottomrule[1pt]
%         & & \multicolumn{2}{c}{Mean Errors}\\
%         Using RGB & Neighbor Rending & Downsample^{1} &  Trans.[m]$\downarrow$ & Rot.[$^{\circ}$]$\downarrow$ \\
%         \hline
%         \ding{51} & & &0.796^{*} & 0.477^{*} \\
%          & \ding{51} &  & 0.133 & 0.247 \\
%          \ding{51} &\ding{51}& \ding{51} & 0.218 & 0.274 \\
%          \ding{51} &\ding{51} & & 0.079 & 0.202\\
%         % RGB & NR\\
%         \toprule[1pt]
%         % \textbf{Ours} & \textbf{4.75} & \textbf{1.16} & \textbf{1.67} \\
%     \multicolumn{4}{l}{$^{\mathrm{*}}$Only five successful scenarios were included in the statistics.}\\
%     \multicolumn{4}{l}{$^{\mathrm{1}}$With a downsampling radius of 10 cm}
%     \end{tabular}}
% \end{table}

\begin{table}[t]
\vspace{1.0em}
    \setlength{\abovecaptionskip}{-0em}
    \centering
    \caption{Ablation Experiments Results}
    \label{Ablation}
    \resizebox{\columnwidth}{!}{
    \begin{tabular}{cc|cc}
        \bottomrule[1pt]
        & & \multicolumn{2}{c}{Mean Errors}\\
        Using RGB & Neighbor Rending &  Trans.(m)$\downarrow$ & Rot.($^{\circ}$)$\downarrow$ \\
        \hline
        \ding{51} & &0.796$^{*}$ & 0.477$^{*}$ \\
         & \ding{51} &  0.133 & 0.247 \\
         % \ding{51} &\ding{51}& \ding{51} & 0.218 & 0.274 \\
         \ding{51} &\ding{51} & \textbf{0.079} & \textbf{0.202}\\
        % RGB & NR\\
        \toprule[1pt]
        % \textbf{Ours} & \textbf{4.75} & \textbf{1.16} & \textbf{1.67} \\
    \multicolumn{4}{l}{$^{\mathrm{*}}$Only five successful scenarios were included in the statistics.}\\
    % \multicolumn{4}{l}{$^{\mathrm{1}}$With a downsampling radius of 10 cm}
    \end{tabular}}
\vspace{-1.0em}
\end{table}

% \begin{CJK*}{UTF8}{gbsn} 
% \lnote{ 现在测试数据的介绍逻辑比较混乱；
% 参考以下逻辑：
% 1.我们在哪些数据集上测试我们的算法（kitti或自己录或其他）；
% 2.数据集的camera设备和lidar设备具体情况（型号，数量等等）；
% 3. 包括多少帧图像和多少帧点云；
% 4.外参真值如何获得的；时间对齐（如有）怎么做的。
% 可参考下面注释的示例：
% }
% \end{CJK*}

% 测试数据介绍示例
% We perform experiments on two datasets. 
% The first dataset is derived from the KITTI Odometry Benchmark~\cite{geiger2013vision}, which includes a Velodyne HDL-64E LiDAR and a high-resolution color camera with a scanning frequency of 10 Hz. 
% The data used for the experiment is synchronized and rectified, and the ground truth for extrinsic parameters can be obtained from the provided calibration files.
% The second one is collected from our autonomous vehicle, equipped with a RoboSense RS-LiDAR-32 LiDAR sensor and a color camera. 
% The dataset, synchronized at a rate of 10 Hz and rectified, consists of 5986 frames containing both images and point clouds. 
% Our acquisition device is shown in the Fig.~\ref{fig:ourdata}.

\subsection{Evaluation Metrics}
\label{Metrics}
% \subsubsection{Extrinsic Parameters}
The assessment of extrinsic parameters involves quantifying the mean translation error across the three axes and mean rotation error for roll, pitch, and yaw. Furthermore, we incorporate ground distance measurements as utilized in roadside camera calibration methods~\cite{vuong2024toward}, to evaluate the accuracy of our extrinsic parameters estimates.
Specifically, we utilize the distance $d_i$ between cones in points cloud as the ground truth and estimate these distances using the corresponding pixel positions, denote as $\Bar{d}_i$. 
The normalized distance measurement error is calculated as $r_i = \frac{|\Bar{d}_i-d_i|}{d_i}$.
It's worth noting that in~\cite{vuong2024toward}, $\Bar{d}_i$ is estimated using the distance of intersection points of rays and ground plane. 
Since we are registering image with point cloud rather than the ground plane, we use the distance of 3D points corresponding to the pixels to estimated $\Bar{d}_i$.
% \begin{CJK*}{UTF8}{gbsn} 
% \lnote{
% 作为独立一小节，一句话实在是太短了。
% 要不扩充具体写计算公式，参考4.2. Evaluation Metrics from LCCNet: LiDAR and Camera Self-Calibration using Cost Volume Network；
% 要不和深度估计合并为一段；
% }
% \end{CJK*}

% Additionally, we utilized the estimated extrinsic parameters to project the 3D vertices of the cones onto the images, calculating the mean reprojection error with corresponding pixels to comprehensively assess the accuracy of the extrinsic parameters.
% \subsubsection{Ground Distance Evaluation}

 \subsection{Results and Analysis}
 
Table~\ref{registration_errors} presents the registration results of our method in eight different scenes, which achieves an average translation error of $0.079m$ and a rotation error of $0.202^{\circ}$. The maximum translation error does not exceed $0.17m$, and the maximum rotation error is around $0.3^{\circ}$, demonstrating the accuracy and robustness of our approach. Fig.~\ref{qualitative} visualizes the registration results of scenes 4 (left) and 5 (middle). 
To replicate errors in manually selecting camera positions in point cloud, we added a $ \pm2m$ perturbation to the displacement vector of pseudo ground truth, treating this as the camera's position for further testing.
Our experiments show that our method robustly handles this variation, automatically estimating initial guess across all tested scenes.

Given that the preprocessing and initial guess estimation processes outlined in~\cite{koide2023general} are ineffective on our dataset, we adopt the SOTA automatic line-based calibration method~\cite{continue_line} as our baseline. 
%读着不顺畅，应当是将调整后的真值作为初值
We add a disturbance($ \pm0.5m$ for translation and $ \pm 1^{\circ}$ for rotation) to ground truth as initial guess for \cite{continue_line}. 
% This perturbation introduces minimal changes to the camera's field of view.
As shown in Table~\ref{registration_errors}, \cite{continue_line} achieves satisfactory registration in some scenes, its performance varies significantly across different scenes due to a scarcity of rich planar features.
It's noteworthy that both our method and \cite{continue_line} demonstrate relatively minimal errors in rotation. 
This is because roadside cameras have a wide observation range, and minor angular deviations may be magnified over distances, making the errors more detectable and more amenable to optimization by the algorithm.

Table~\ref{Distance Evaluation} displays the performance of our method in ground distance measurement on the self-collected dataset. 
We evaluate from the perspectives of maximum error, median error, and root-mean-squared error (RMSE, in $\%$). 
Compared to SOTA method~\cite{vuong2024toward}, our distance measurement results show significant improvement across all evaluation metrics, indicating the precision of the registration method.

% As analyzed in Section~\ref{Initial Guess Estimating}, our methodology leverages the abundant and uniformly distributed features in images, alongside environments that are favorable for point cloud acquisition. 
% However, these requirements are not met in all scenarios. 
In extreme conditions where ground texture features are minimal and excessive tree density obstructs drones from capturing point clouds effectively, our method may fail.
In Fig.~\ref{bad scenes}, we illustrate scenarios where our method fails to function properly. 
In the depicted left scene, the road surface lacks sufficient texture features and the absence of features in the vicinity making it difficult to obtain enough feature points for matching.
The image on the right shows the camera's left FoV being obstructed, with features predominantly centralized. 
Moreover, the dense foliage of trees complicates the acquisition of point clouds, resulting in substantial registration inaccuracies. 

\subsection{Ablation studies}

We investigated the impact of \emph{neighbor rendering} module and appearance of point clouds to the overall method.
Table~\ref{Ablation} presents our results.
Using intensity information instead of RGB values only results in a slight decrease in registration accuracy, thanks to SuperGlue's robust capability in matching heterogeneous images.
Omitting \emph{neighbor rendering} from image generating resulted in successful registration in just five of eight scenarios, accompanied by a significant drop in accuracy.
% Success was mainly in open, expansive environments. 
This issue arises from distant points in the upper part of the image forming dense pixels, whereas closer points at the bottom don't without \emph{neighboring rendering}, causing match concentration in the upper area and biasing extrinsic parameter estimation towards a local optimum.

To evaluate the impact of point cloud density on different rendering methods, we conducted tests assessing their sensitivity to density variations. 
We observe that direct projection struggles to produce matchable images across all scenes when the point cloud is downsampled by a 3 cm radius. 
Conversely, incorporating \emph{neighbor rendering} allows the method to function effectively, even with a 8 cm downsampling radius, maintaining registration precision.
Post-downsampling registration outcomes are depicted to the right of Fig.~\ref{qualitative}, with rendering details in the appendix.

\section{Applications}

We apply the registration framework to roadside 3D object detection.
We initially employ YOLOv7~\cite{wang2023yolov7} for 2D object detection, estimating the bottom centers of detection boxes as the interface points between vehicles' rear ends and the ground. 
Using the pixel coordinates of these points, we can retrieve the corresponding 3D points from the registered point cloud, which serves as the 3D position of the target for 3D object detection. 
% Fig.~\ref{application_device} is the visualization of 3D object detection results.
% \begin{figure}[t]
% \vspace{0.2cm}
% \begin{center}
% 	\begin{minipage}{1\linewidth}
%         {\includegraphics[width=0.49\textwidth]{fig/GPS小车.jpg}}
%         {\includegraphics[width=0.49\textwidth]{fig/3D_DET.jpg}}
% 	\end{minipage}
% 	\caption{Left: Vehicle equipped with GPS-RTK to provide ground truth. Right: visualization of 3D object detection results}
% \label{application_device}
% \end{center}
% \vspace{-1.0em}
% \end{figure}

\begin{figure}[t]
\vspace{0.2cm}
\begin{center}
	\begin{minipage}{1\linewidth}
        {\includegraphics[width=1\textwidth]{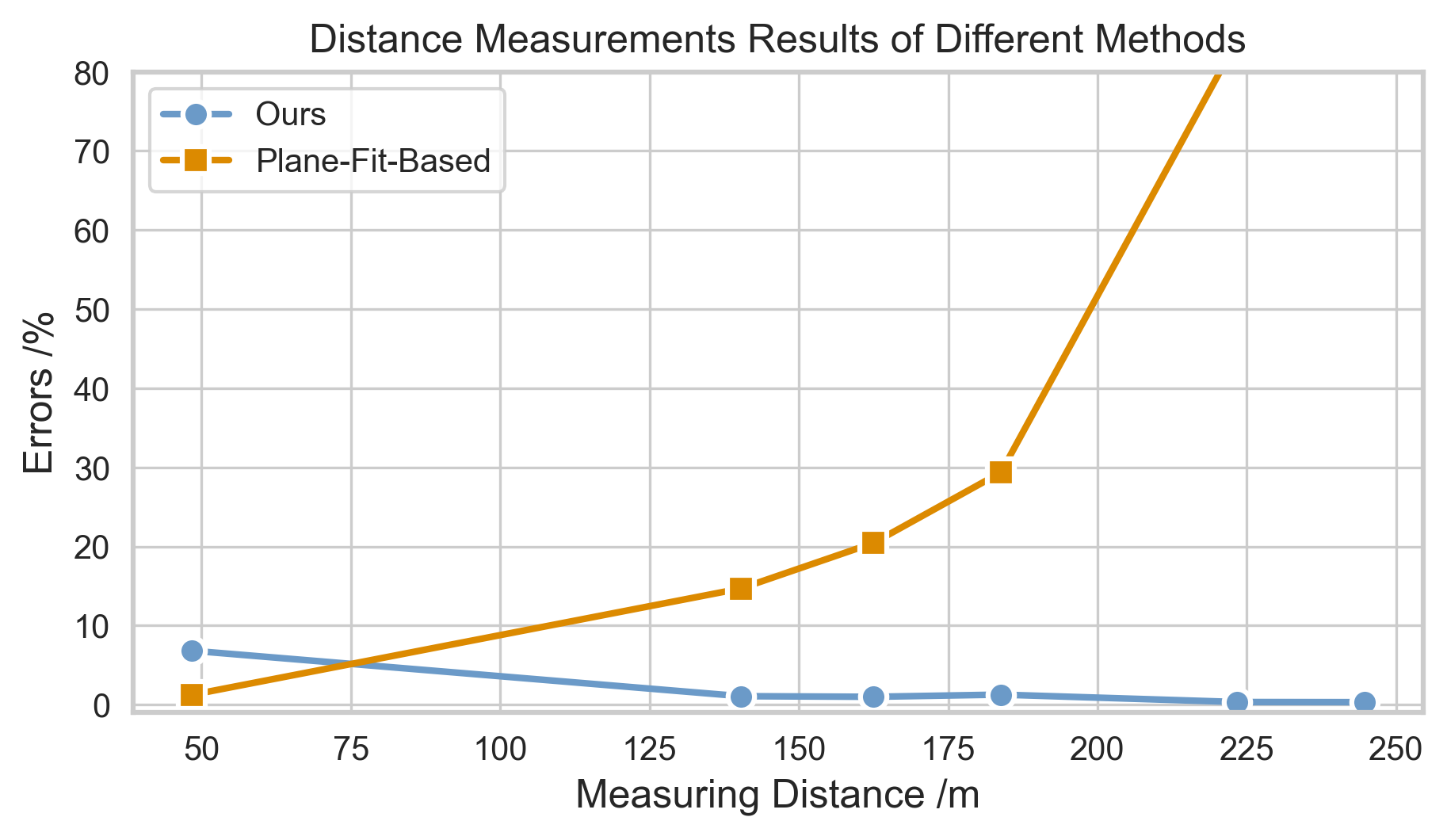}}
	\end{minipage}
	\caption{Comparing distance measurements from different methods, the ground truth is provided by the GPS-RTK equipped on the vehicle.}
\label{distance measurements}
\end{center}
\vspace{-1.0em}
\end{figure}

We utilize a GPS-RTK-equipped vehicle, time-synchronized with the roadside camera, to provide positional ground truth. 
The distance from the GPS device to the vehicle's rear is manually gauged.
Considering the limitation of GPS to only two-dimensional (x and y coordinates) data, we simplify the evaluation by transforming 3D detection positions into distances from the target to the camera, adopting the distance measurement error from Section~\ref{Metrics} as our evaluative metric.
Owing to spatial constraints, details on evaluation equipment and object detection result visualizations are provided in the appendix.

Fig.~\ref{distance measurements} depicts the distance measurements results of our method.
Within a 250-meter detection range, most of the errors between our measurements and the ground truth is controlled within approximately $1.5\%$.
The notable error at 50 meters is due to approximation error arising from using the 2D bbox’s bottom center to estimate the vehicle’s rear end ground contact point at close distances, but this approximation error decreasing as the distance increases.
Road surface irregularities cause fitting plane methods to incur substantial errors, which increase with distance. 
Conversely, the registered point cloud effectively mitigates this issue.
This results demonstrates the significance of fusing prior point clouds with images for visual-based detection tasks and also proves the precision of our registration method.

\section{Conclusion}
This paper presents an automated approach for registering prior point clouds with roadside camera images. 
We introduce an efficient rendering method aimed at minimizing discrepancies between generated images and actual camera photos to enhance registration. 
Moreover, we outline a workflow capable of precisely estimating initial guess using merely approximate camera locations, markedly decreasing the registration process's dependence on initial configurations. 
We achieving a rotation accuracy of 0.202$^{\circ}$ and a translation precision of 0.079m in self-collected dataset.

In the future, we plan to investigate our methodology's applicability in varied contexts, including indoor scenes and tunnels, and to enhance performance by combining our approach with advanced rendering methods.

\section*{APPENDIX}
\setcounter{figure}{0}
\renewcommand{\thefigure}{A\arabic{figure}}

\subsection{Visualization of Neighbor Rendering}
\begin{figure}[th]
\vspace{0.2cm}
\begin{center}
    \subfigure{
        \rotatebox{90}{\scriptsize{~~~~~~~Without NR}}
            \begin{minipage}[t]{0.44\linewidth}
                \centering
                \includegraphics[width=1\linewidth]{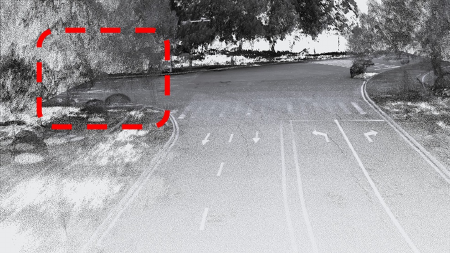}
            \end{minipage}
    	}
     \subfigure{
            \begin{minipage}[t]{0.44\linewidth}
                \centering
                \includegraphics[width=1\linewidth]{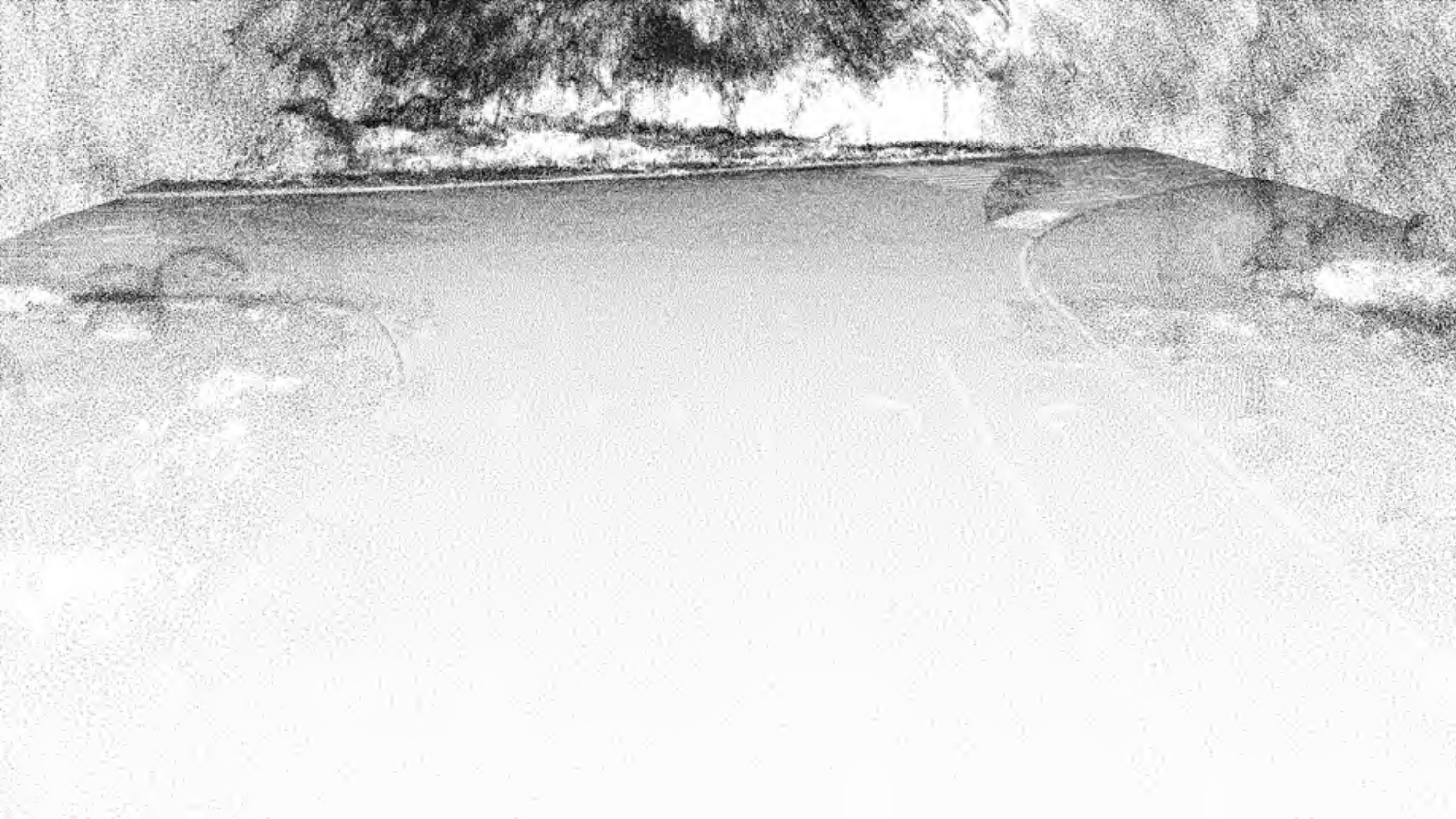}
            \end{minipage}
    	}
     \subfigure{
        \rotatebox{90}{\scriptsize{~~~~~~~With NR}}
            \begin{minipage}[t]{0.44\linewidth}
                \centering
                \includegraphics[width=1\linewidth]{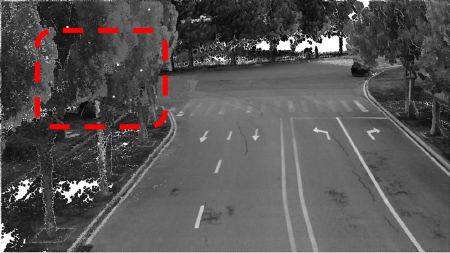}
                \centerline{Radius=$3cm$}
            \end{minipage}
    	}
     \subfigure{
            \begin{minipage}[t]{0.44\linewidth}
                \centering
                \includegraphics[width=1\linewidth]{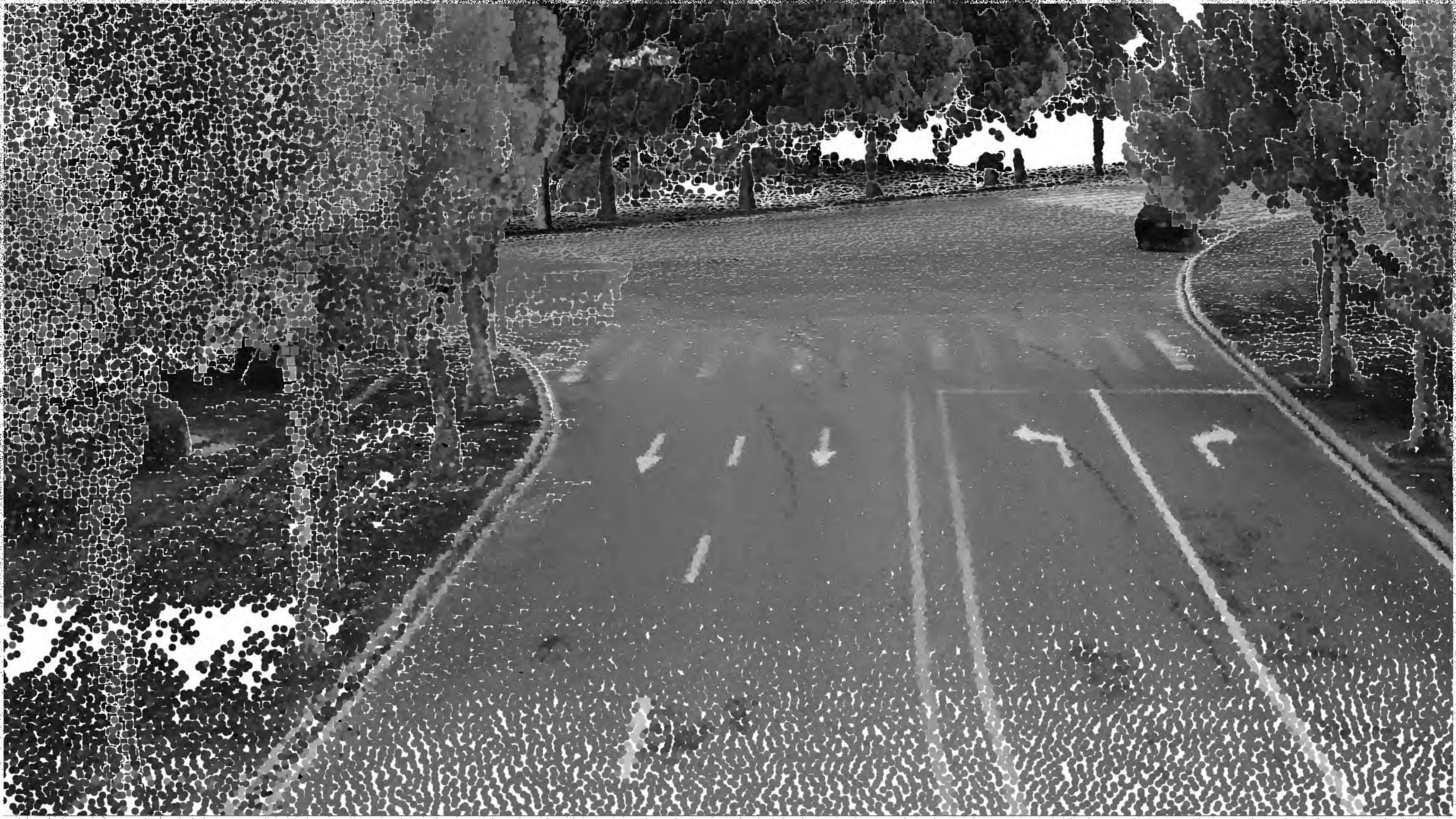}
                \centerline{Radius=$8cm$}
            \end{minipage}
    	}
	\caption{The top row displays images generated by directly projecting the point cloud and filtering occluded points, while the bottom row shows images generated using \emph{neighbor rendering (NR)}. Utilize RGB information to generate grayscale values.}
\label{render_comp}
\end{center}
\vspace{-1.0em}
\end{figure}

Fig.~\ref{render_comp} demonstrates the effects of different rendering methods on point clouds of varying densities. 
It is evident that when the point cloud downsampling radius is $3cm$, the method of direct projection struggles to form clear images, with white holes occupying a significant portion.
Additionally, within the red box, road surfaces that should be obscured by trees are visible due to these holes, a phenomenon known as the bleeding problem.
Employing \emph{neighbor rendering} effectively addresses these issues, creating views as realistic as the original camera images. 
At an $8cm$ downsampling radius, directly projecting the point cloud fails to convey any meaningful information, whereas neighbor rendering still maintains clear and visible textures, effectively mitigating the bleeding problem.

\subsection{Experiments Results and Scenes Visualization}

\begin{figure}[th]
\vspace{0.2cm}
\begin{center}
	\begin{minipage}{1\linewidth}
        \centering
        {\includegraphics[width=0.49\textwidth]{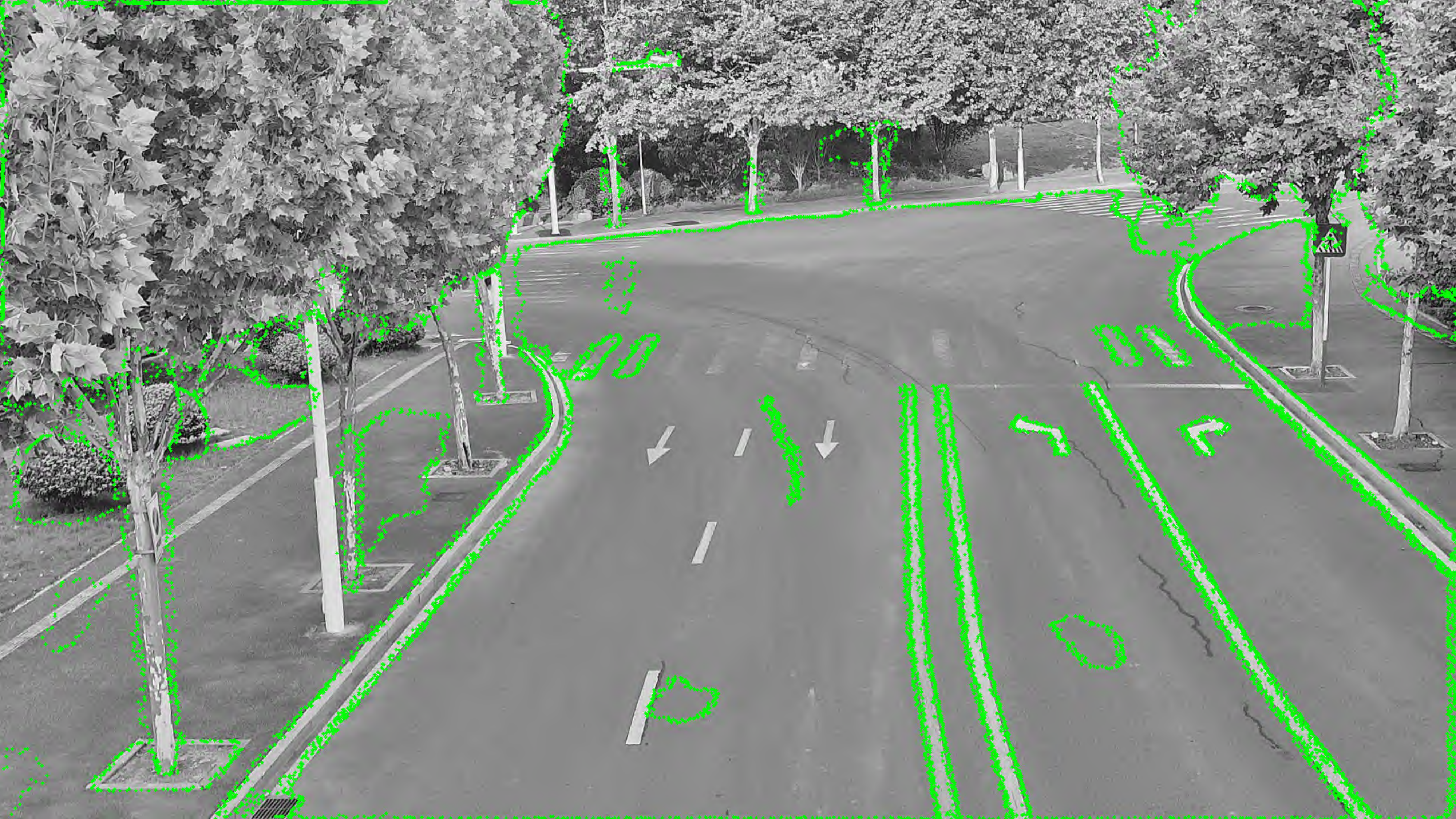}}
        {\includegraphics[width=0.49\textwidth]{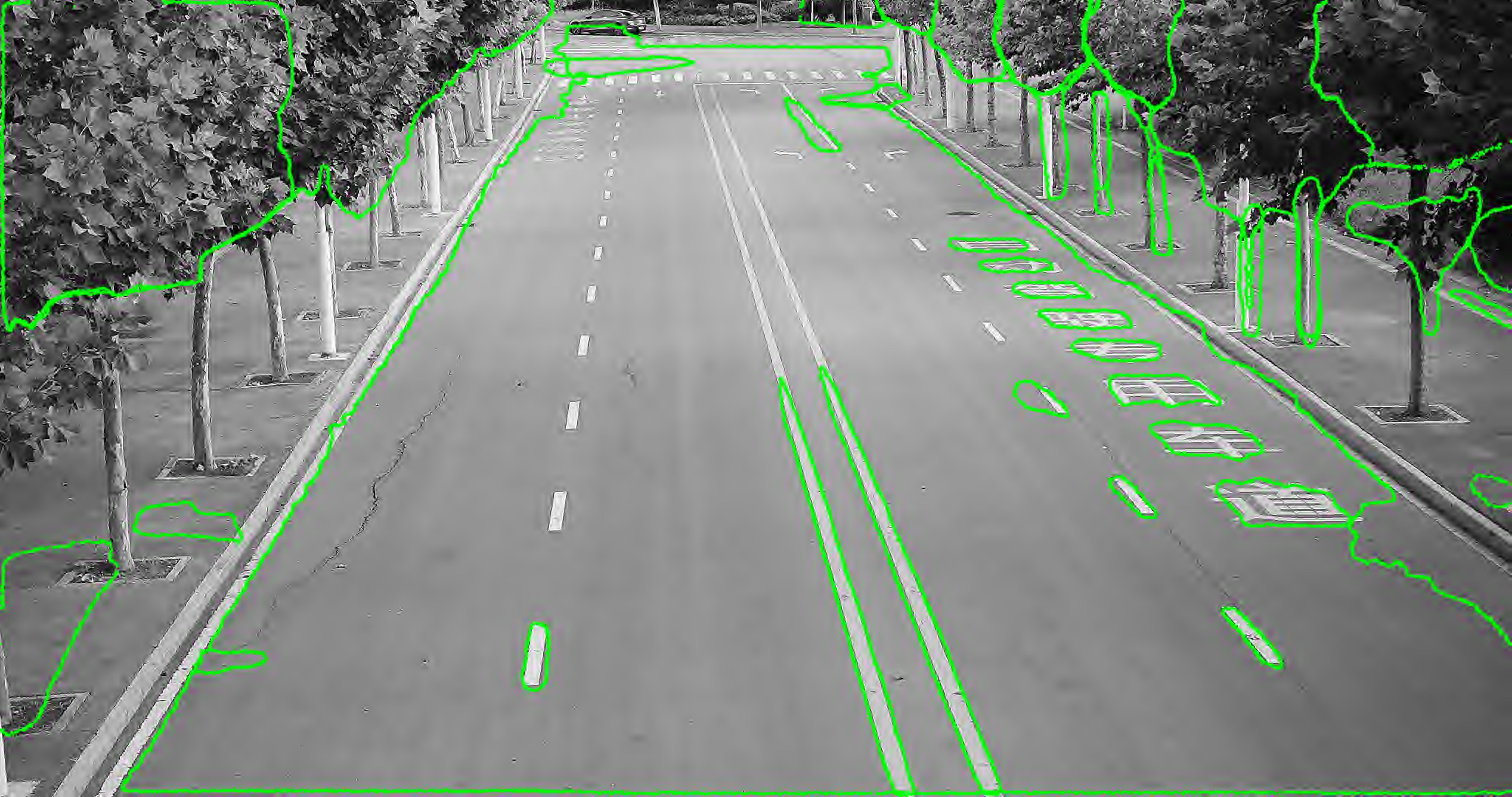}}
        {\includegraphics[width=0.49\textwidth]{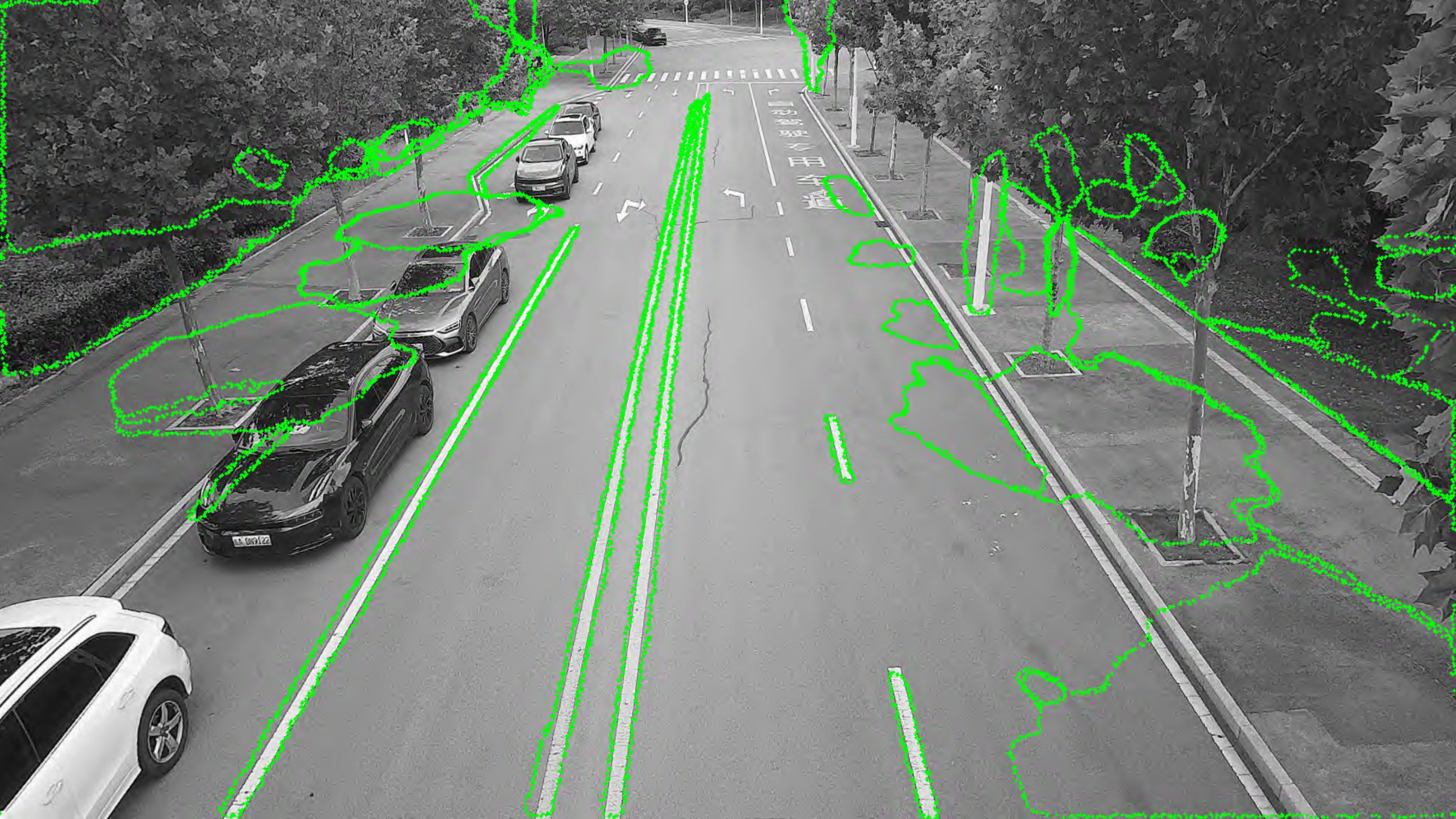}}
        {\includegraphics[width=0.49\textwidth]{fig/pdf/quantative_2.pdf}}
        {\includegraphics[width=0.49\textwidth]{fig/pdf/quantative_1.pdf}}
        {\includegraphics[width=0.49\textwidth]{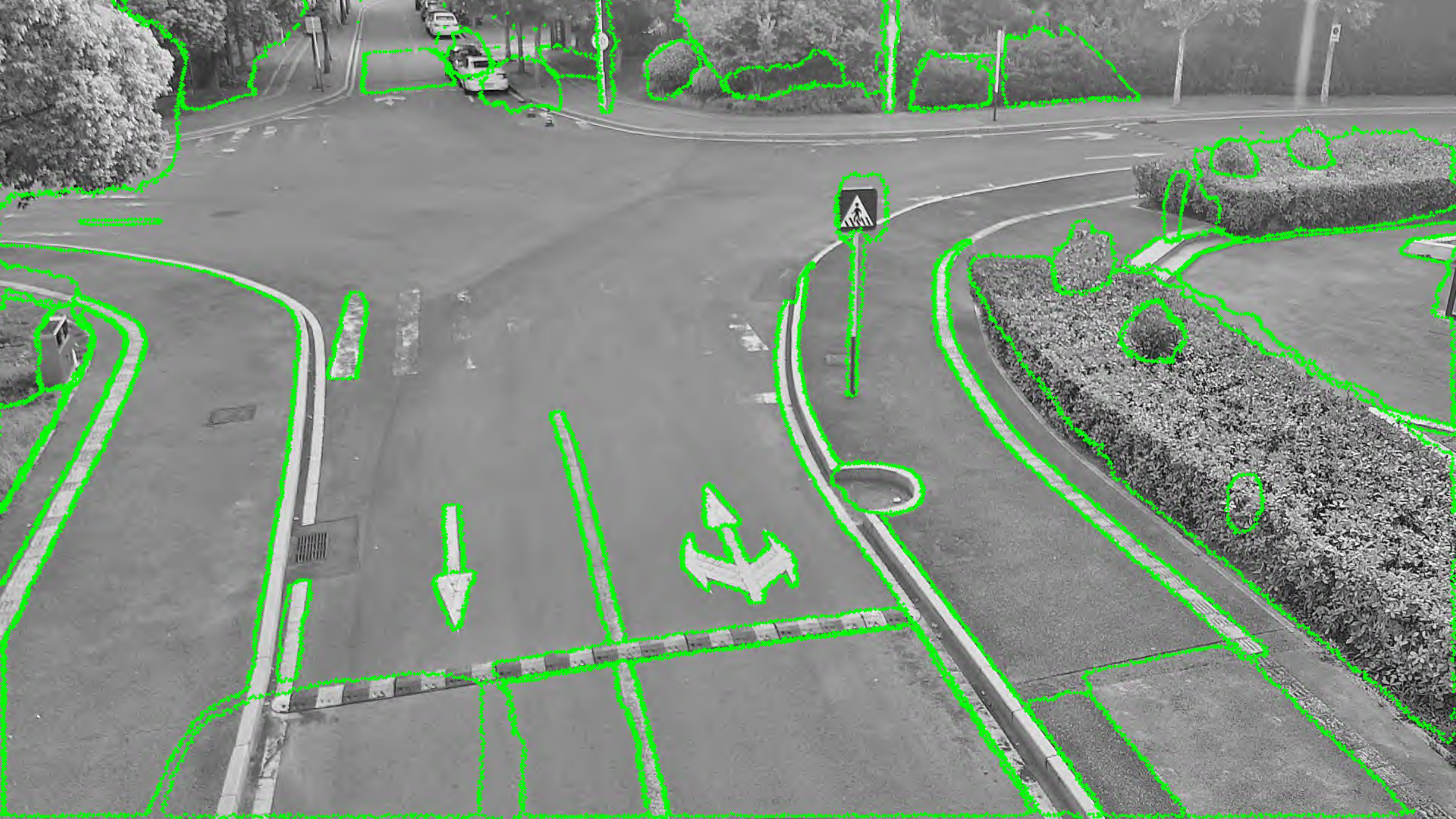}}
        {\includegraphics[width=0.49\textwidth]{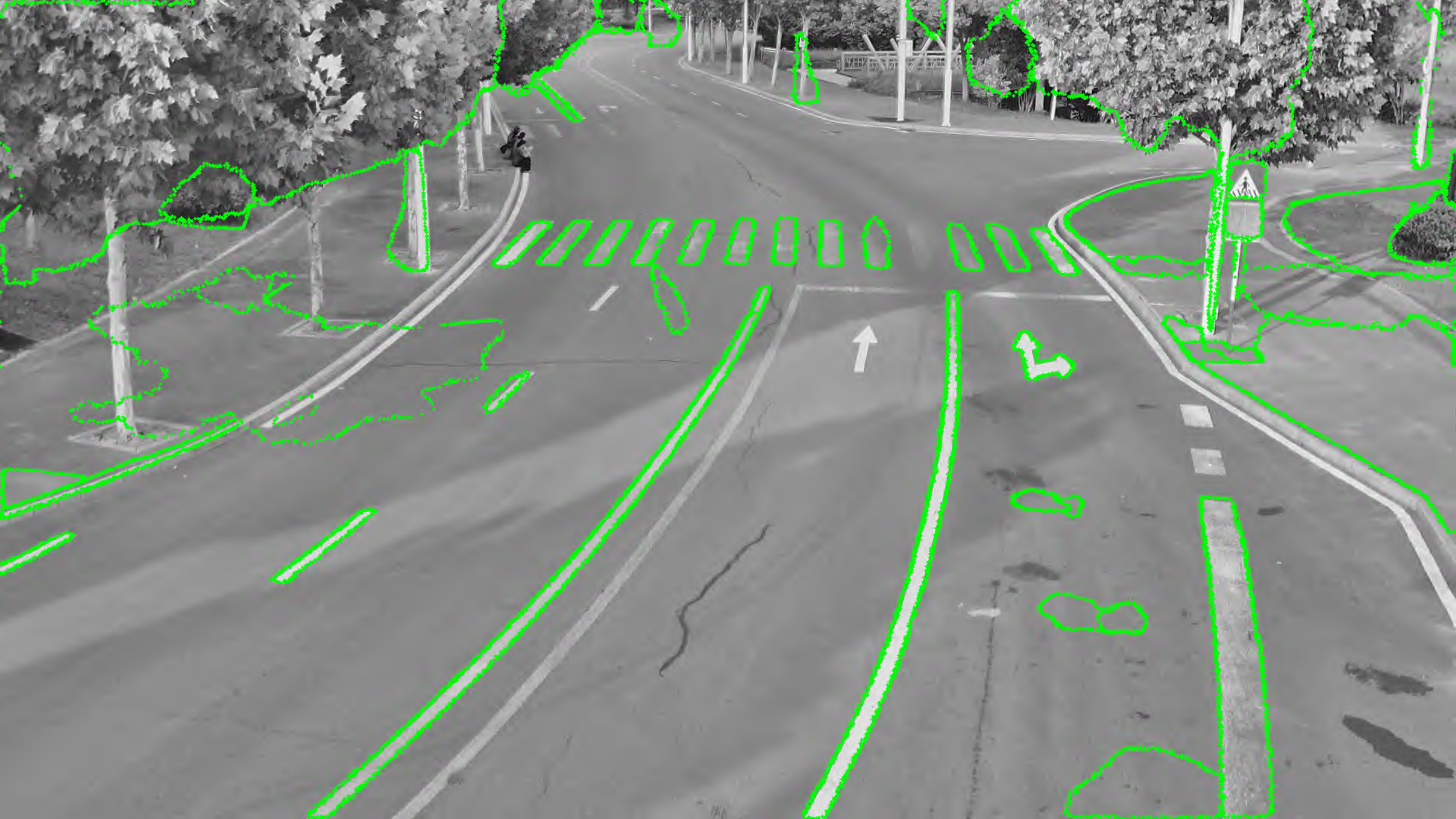}}
        {\includegraphics[width=0.49\textwidth]{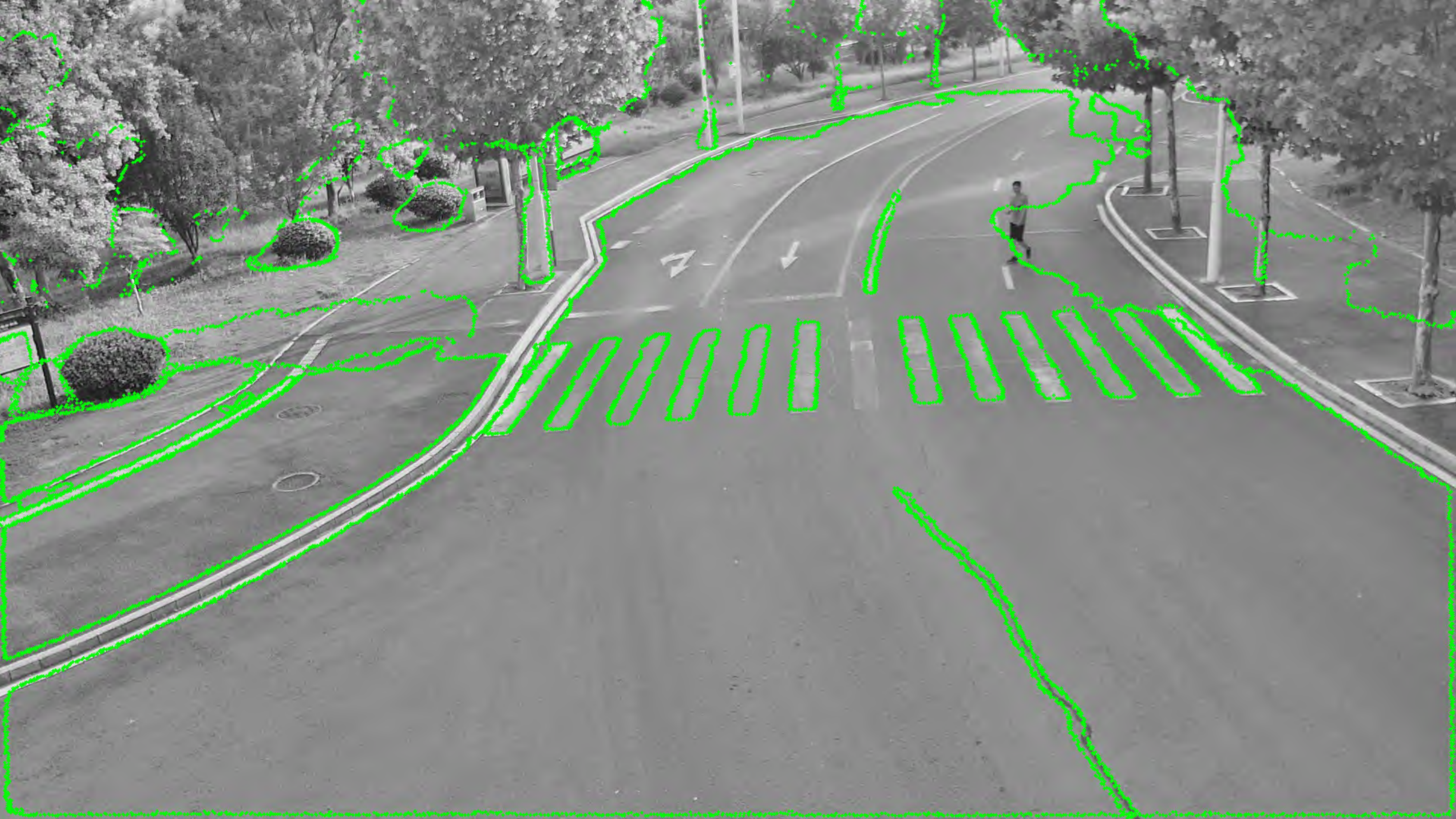}}
	\end{minipage}
	\caption{From left to right, and top to bottom, are the visualizations of scenes 1 to 8, in sequence. The green lines represent the results of projecting the edges from the point cloud onto the image using the registered extrinsic parameters.}
\label{exp_results}
\end{center}
\vspace{-1.0em}
\end{figure}

In Fig.~\ref{exp_results}, we present the scenes utilized in our experiments alongside the visualizations of their registration outcomes. 
Our algorithm consistently achieve accurate registration across all eight scenes. 
Specifically, in Scene 3, the inability of several green circles on the left to align with image objects stems from dense foliage obstructing the drone's capacity to capture point clouds under the trees. 
 Nevertheless, our technique adeptly registers point clouds with images, highlighting the resilience and effectiveness of our methodology.

\subsection{3D Object Detection Application}

\begin{figure}[th]
\vspace{0.2cm}
\begin{center}
	\begin{minipage}{1\linewidth}
        {\includegraphics[width=0.49\textwidth]{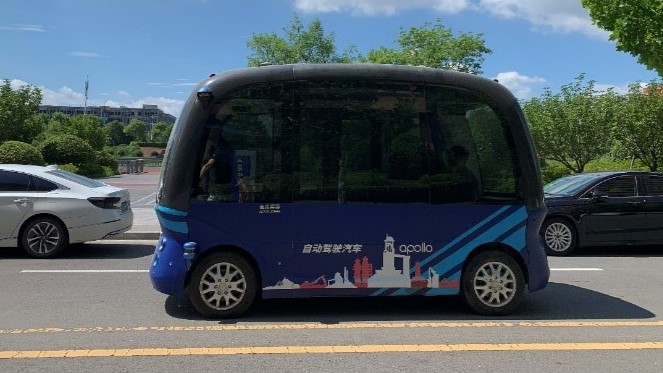}}
        {\includegraphics[width=0.49\textwidth]{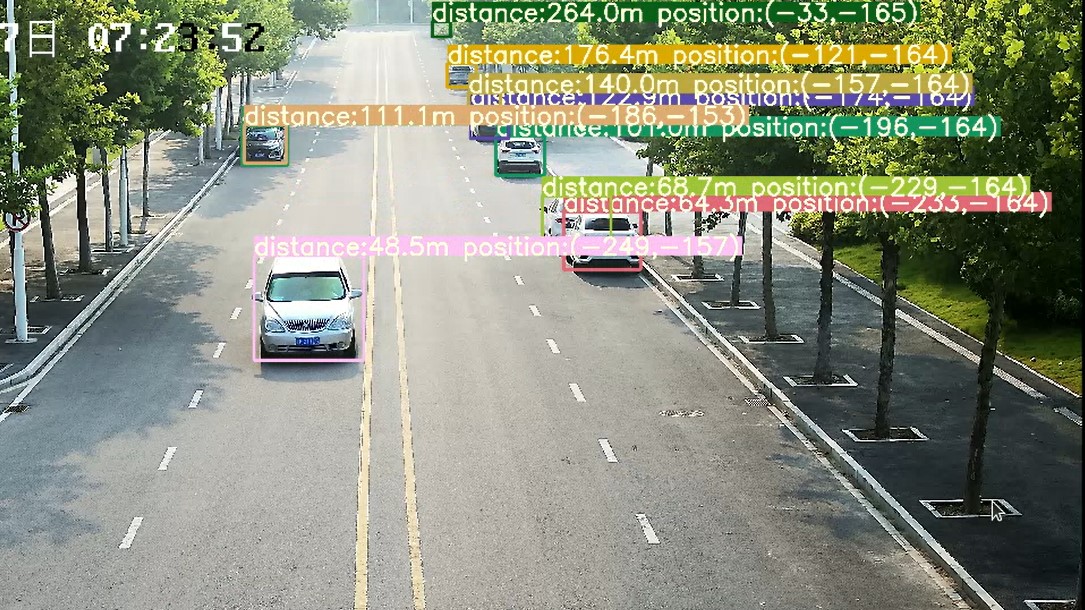}}
	\end{minipage}
	\caption{Left: Vehicle equipped with GPS-RTK to provide ground truth. Right: visualization of 3D object detection results}
\label{application_device}
\end{center}
\vspace{-1.0em}
\end{figure}

We utilize a vehicle outfitted with GPS-RTK to obtain 3D positional ground truth. The visualization of 3D object detection outcomes is shown in the right panel of Fig~\ref{application_device}. 
Given that GPS only yields $xy$ coordinates as ground truth, our analysis is conducted from a Bird's Eye View (BEV), concentrating exclusively on $xy$ results and omitting $z$ values.

We identify the contact point between vehicles and the ground using the center of the bottom edge of the green 2D detection boxes depicted in the right side of Fig.~\ref{application_device}.
The 3D point in the registered point cloud corresponding to this pixel is designated as the target's 3D location.
The two-dimensional coordinates shown in  Fig.~\ref{application_device} indicate the $xy$ position of this 3D point.

% \bibliographystyle{IEEEtran}
% \bibliography{IEEEabrv,my}

% \balance

%  \begin{appendices}  
% \setcounter{figure}{0}
% \renewcommand{\thefigure}{A\arabic{figure}}
% \input{component/appendix}
% \end{appendices}  

\end{document}